%% file: article.tex
\documentclass[a4paper, 10pt, authoryear, sort, 3p, final]{elsarticle}

\usepackage[utf8]{inputenc}
\usepackage[T1]{fontenc}
\usepackage{amsfonts}
\usepackage{amsmath}
\usepackage[safe]{tipa}
\DeclareUnicodeCharacter{00E7}{\c{c}} 
\DeclareUnicodeCharacter{014B}{\textipa N} 
\DeclareUnicodeCharacter{0251}{\textscripta} 
\DeclareUnicodeCharacter{0254}{\textopeno} 
\DeclareUnicodeCharacter{025B}{\textepsilon} 
\DeclareUnicodeCharacter{0259}{\textschwa} 
\DeclareUnicodeCharacter{0261}{\textscriptg} 
\DeclareUnicodeCharacter{0281}{\textinvscr} 
\DeclareUnicodeCharacter{0283}{\textesh} 
\DeclareUnicodeCharacter{0289}{\textbaru} 
\DeclareUnicodeCharacter{028C}{\textturnv} 
\usepackage{mathptmx}
\usepackage{relsize}

\usepackage{titlesec}

\usepackage{graphicx}

\usepackage{subcaption}
\newcommand*{\eg}{e.g.}
\newcommand*{\ie}{i.e.}

\usepackage[single]{acro}
\DeclareAcronym{fps}{short=fps, long=frames per second}
\DeclareAcronym{ncc}{short=NCC, long=normalized cross-correlation}
\DeclareAcronym{dof}{short=DoF, long=degrees of freedom}
\DeclareAcronym{dfki}{short=DFKI, long=German Research Center for Artificial Intelligence}
\DeclareAcronym{ema}{short=EMA, long=electromagnetic articulography}
\DeclareAcronym{us}{short=US, long=ultrasound}
\DeclareAcronym{epg}{short=EPG, long=electropalatography}
\DeclareAcronym{cr}{short=CR, long=cineradiography}
\DeclareAcronym{ct}{short=CT, long=computed tomography}
\DeclareAcronym{cbct}{short=CBCT, long=cone beam computed tomography}
\DeclareAcronym{fov}{short=FOV, long=field of view}
\DeclareAcronym{hosvd}{short=HOSVD, long=higher-order singular value decomposition}
\DeclareAcronym{ipa}{short=IPA, long=international phonetic alphabet}
\DeclareAcronym{mri}{short=MRI, long=magnetic resonance imaging}
\DeclareAcronym{rtmri}{short=rtMRI, long=real-time magnetic resonance imaging}
\DeclareAcronym{nmr}{short=NMR, long=nuclear magnetic resonance}
\DeclareAcronym{pde}{short=PDE, long=partial differential equation}
\DeclareAcronym{pca}{short=PCA, long=principal component analysis}
\DeclareAcronym{roi}{short=ROI, long=region of interest}
\DeclareAcronym{lca}{short=LCA, long=linear component analysis}
\DeclareAcronym{capt}{short=CAPT, long=computer-aided pronunciation training}

\usepackage[obeyFinal, colorinlistoftodos]{todonotes}
\usepackage[hidelinks]{hyperref}

\usepackage[all]{hypcap}
\usepackage{doi}

\usepackage{siunitx}
\renewcommand\vec{\mathbf}

\usepackage{tabulary}
\usepackage{booktabs}

\makeatletter
\def\ps@pprintTitle{%
 \let\@oddhead\@empty
 \let\@evenhead\@empty
 \def\@oddfoot{}%
 \let\@evenfoot\@oddfoot}
\makeatother

\begin{document}

\begin{frontmatter}

\title{A Multilinear Tongue Model Derived from Speech Related MRI Data of the Human Vocal Tract}

\author[mmci,dfki,sgcs]{Alexander Hewer\corref{cor1}}
\ead{hewer@coli.uni-saarland.de}
\author[inria]{Stefanie Wuhrer}
\ead{stefanie.wuhrer@inria.fr}
\author[mmci,dfki]{Ingmar Steiner}
\ead{steiner@coli.uni-saarland.de}
\author[cstr]{Korin Richmond}
\ead{korin@cstr.ed.ac.uk}

\cortext[cor1]{Corresponding author}

\address[mmci]{Cluster of Excellence ``Multimodal Computing and Interaction'',
  Saarland University, Saarbrücken, Germany}
\address[dfki]{German Research Center for Artificial Intelligence (DFKI GmbH),
  Saarbrücken, Germany}
\address[sgcs]{Saarbrücken Graduate School of Computer Science,
  Saarbrücken, Germany}
\address[inria]{INRIA Grenoble Rhône-Alpes, France}
\address[cstr]{Centre for Speech Technology Research,
  University of Edinburgh, UK}

\begin{abstract}
  \input{abstract}
\end{abstract}


\end{frontmatter}

\section{Introduction}

\subsection{Motivation}

As one of the main articulators, the tongue plays a major role in human speech production.
In speech science, it is therefore of great interest to understand its shape and motion during speech articulation.
To this end, we want to derive a tongue model that offers the following features:
it is a three-dimensional articulatory model that only uses a few \ac{dof} that influence the shape of the tongue.
Furthermore, the \ac{dof} of the model are split into two sets:
One set controls the anatomical shape of the tongue and the other, the speech related tongue pose.
This is important because the articulation strategy may depend on the anatomy of the speaker \citep{johnson1993individual, ladefoged1957information, honda1996human, brunner2009relationship, fuchs2008speakers, rudy2013effect, weirich2013palatal, weirich2013inter, yunusova2012positional}.
As shape representation of the model, we select a polygon mesh that can easily be used in various fields of applications:
in computer graphics, such meshes are used to generate animations of complex objects \citep{Botsch2010} or to model objects of highly complex geometry and topology.
Additionally, polygon models have been used in speech processing to generate acoustical simulations \citep{Blandin2015JASA}.

The model itself can be used to equip virtual avatars for multimodal spoken interaction with a more natural animation of the tongue.
In this regard, we note that it is of vital importance to synthesize the correct motion for the speech audio:
\citet{Mcgurk1976} found that inconsistencies between visible mouth motions and audible speech may cause the speech to be perceived incorrectly.
Moreover, information about tongue motion can be applied in \ac{capt} to provide the user with visual information on how to move the tongue to produce a specific sound \citep{Engwall2008}.
It can also be employed in an articulatory speech synthesis framework to help approximate the vocal tract area function or it can be used to estimate the full tongue shape from sparse data, such as \ac{ema} measurements.
Finally, the model can also help to perform speaker normalization, that is, investigate only shape variations of an articulation that is independent of the speaker anatomy.
In particular, such a model allows speaker adaptation, which is useful in the aforementioned areas of applications.
For example, in audiovisual speech synthesis, \ac{capt}, and articulatory speech synthesis, it is vital to replicate the speaker's specific tongue shape to match the remaining anatomy;
the tongue should not leave the mouth or penetrate the palate during articulation.
Additionally, for \ac{capt}, the speaker's tongue anatomy influences the articulatory strategy of the speaker.
Providing the incorrect strategy could confuse the subject, especially if real-time feedback was provided from \ac{ema} data.
In the case of estimating the tongue shape from \ac{ema}, using the wrong anatomy of the tongue may keep the model from registering the \ac{ema} data correctly.

For completeness, we want to mention another class of tongue models, so-called biomechanical models.
Such models aim at simulating the entire tongue body, including the internal muscle activities.
This property is useful for, \eg, simulating laryngoscopy \citep{rodrigues2001biomechanical}, investigating the consequences of surgery \citep{buchaillard2007simulations}, or for studying muscle activation during speech \citep{buchaillard2009biomechanical, wu2014iterative}.
However, for our target application areas, such a model may be regarded as too complex.

\subsection{Methodology}

We want to derive \ac{dof} that are speech related.
Thus, we need to analyze meshes extracted from actual speech production data.
However, most of the articulators are contained inside the human mouth and therefore partially or completely hidden from view.
This means that traditional imaging modalities based on light, \eg, photography, are of limited use for acquiring the desired shape information for analysis.
Currently, \ac{mri} can be regarded as the state-of-the-art technique for investigating the interior of the human vocal tract during speech.
It is non-invasive and non-hazardous to the subject, and in contrast to \ac{us} or \ac{ema}, it is able to provide dense volumetric measurements.
Moreover, a lot of previous work has focused on adapting the \ac{mri} method to the needs of speech research.
The main issue in early studies \citep{baer1991analysis} was the long acquisition time, which forced subjects to maintain the vocal tract configuration for a long time with brief interruptions:
one scan took around 3 minutes.

Subsequent advances in \ac{mri} scanners made it possible to acquire 3D time-evolving models of the vocal tract \citep{foldvik1993time, shadle1999} and 2D \ac{mri} movies with up to 5 \ac{fps} \citep{demolin2000real}.
An approach to \ac{rtmri} recording of the vocal tract with synchronized audio was presented by \citet{narayanan2004}, which offered a frame rate of up to 24 \ac{fps} and thus enabled the examination of the dynamics of fluent speech using \ac{mri}.
This method also applied noise cancellation to deal with the scanner noise.
More recent methods \citep{Kim2009, scott2012, niebergall2013real, burdumy2015acceleration, fu2015high, elie2016, Lingala2016, lingala2017fast} further reduced the acquisition time and improved the quality of obtained scans.
For example, \citet{lingala2017fast} reported \ac{rtmri} scanning at 83 \ac{fps} for a single slice, or 27 \ac{fps} for three slices.

As a lot of speech production data can nowadays be obtained by means of \ac{mri} techniques in a short amount of time, the mesh extraction process from the individual scans should be as minimally supervised as possible, as doing it manually takes a lot of time and typically requires anatomical expertise.
For the analysis of extracted meshes, statistical methods like \ac{pca} can be used.
Such a method provides access to a statistical model that uses a low-dimensional subspace to represent the shape of the corresponding object, and is also able to estimate the plausibility of such a shape.
This approach requires a labeled database that shows the object under consideration in many different shapes that are related to the motion to be observed.
Polygon meshes already have been successfully used for statistical shape analysis of, e.g., human bodies \citep{Allen2003}, faces \citep{Blanz1999}, or tongues \citep{Badin2006}.
As we want to separate the anatomy related variations from the ones related to the speech related tongue pose,
we use a method that leads to a so-called multilinear model.
This statistical model type is able capture those different types of variation separately.

\subsection{Related work}
\acreset{rtmri, dof}

A sizable body of research has focused on analyzing the vocal tract shape during speech production using different modalities,
including X-ray, \ac{epg}, \ac{cr}, \acl{us}, \ac{cbct}, \ac{rtmri}, or \ac{ct}.
In \autoref{tab:related-work-overview}, we provide an overview of previous studies. 

\begin{table}
  \centering
  \smaller
  \begin{tabulary}{\linewidth}{LCLS[table-format=2.0]}
    \toprule
    Work & Modality & Analyzed Data & {Subjects} \\
    \midrule
    \citet{mermelstein1973articulatory} & X-ray & 2D contours & 1 \\
    \citet{Harshman1977} & X-ray & 2D contours & 5 \\
    \citet{baer1991analysis} & \ac{mri} & vocal tract area functions and shapes & 4 \\
    \citet{stone1992representing} & \ac{us} & fitted polynomial functions & 1 \\
    \citet{narayanan1995articulatory} & \ac{mri} & shapes & 4 \\
    \citet{stone1996three} & 3D \ac{us} + \ac{epg} & interpolated meshes & 1 \\
    \citet{tiede1996shape} & \ac{mri} & cross-section shapes & 1 \\
    \citet{narayanan1997toward} & \ac{mri} + \ac{epg} & shapes & 4 \\
    \citet{badin1998three} & \ac{mri} + \ac{cr} & meshes & 1 \\
    \citet{engwall1999collecting} & \ac{mri} & meshes + 2D contours & 1 \\
    \citet{Engwall2000} & \ac{mri} & meshes & 1 \\
    \citet{Hoole2000} & \ac{mri} & 2D contours & 9 \\
    \citet{kroger2000estimation} & \ac{mri} & vocal tract area functions & 1\\
    \citet{beautemps2001linear} & \ac{cr} + labio-film & 2D contours & 1 \\
    \citet{badin2002three} & \ac{mri} + video & meshes & 1 \\
    \citet{Zheng2003} & \ac{mri} & sparse 3D point clouds & 5 \\
    \citet{Badin2006} & \ac{mri} + \ac{ct} & meshes & 1 \\
    \citet{geng2009stretch} & \ac{ema} & flesh points & 7 \\
    \citet{Ananthakrishnan2010} & \ac{mri} & 2D contours & 3 \\
    \citet{Valdes2012} & \ac{mri} & 2D contours & 7 \\
    \citet{toutios2015factor} & \ac{rtmri} & 2D contours & 1 \\
    \citet{kaburagi2015morphological} & \ac{mri} & vocal tract area functions & 10 \\
    \citet{woo2015construction} & dynamic \ac{mri} & images & 18 \\
    \citet{woo2015high} & \ac{mri} & deformation fields & 20 \\
    \citet{stone2016structure} & \ac{mri} & muscle architectures & 14 \\
    \citet{Fang2016} & \ac{mri} + \ac{cbct} & meshes & 1 \\
    \citet{serrurier2017inter} & \ac{mri} & 2D contours & 11 \\
    \bottomrule
  \end{tabulary}
  \caption{Overview of several studies that have investigated shape variabilities of the vocal tract.
    We list the modality (or modalities) used, the analyzed data representation, and the number of subjects taking part in the corresponding study.}
  \label{tab:related-work-overview}
\end{table}

Even some of the earliest studies aimed at analyzing the anatomical and speech related shape variations by using multiple subjects;
\citet{Harshman1977} investigated these variations in 2D X-ray data.
Nowadays, this imaging modality is no longer used for this purpose, due to the dangers of the ionizing radiation involved.
\citet{narayanan1995articulatory, narayanan1997toward} analyzed shape variabilities using 3D \ac{mri} data.
Analysis on 2D \ac{mri} was conducted by \citet{Hoole2000}, \citet{Ananthakrishnan2010}, and \citet{Valdes2012, Vargas2012}.
\citet{Zheng2003} performed this analysis on sparse sets of 65 points that were manually extracted from 3D \ac{mri} scans.
\citet{kaburagi2015morphological} used \ac{pca} to analyze the vocal tract area functions of ten speakers obtained from \ac{mri}.
The work by \citet{woo2015construction} used dynamic \ac{mri} to build a spatio-temporal atlas of the vocal tract. 
\citet{woo2015high} analyzed a high resolution atlas of the vocal tract using \ac{pca}.
In the study by \citet{stone2016structure}, the muscle architectures of different subjects were investigated.
Speaker normalization was performed by \citet{geng2009stretch} and \cite{serrurier2017inter}.

Shape variations related to the anatomy of the subject are also of interest in the field of biomechanical models:
\citet{bijar2016atlas} presented an atlas-based automatic approach to generate subject-specific finite element tongue meshes.
\citet{harandi2017} used cine \ac{mri} to derive speaker-specific biomechanical models.

For our purposes, we need to analyze the anatomical and speech related variations in 3D meshes.
Initial work investigating these variations obtained from \ac{mri} data of 9 speakers was presented by \citet{Hoole2003}, but neither evaluated nor published \citetext{Hoole, personal communication}.
Moreover, work that focused on the speech related shape variations of a more dense 3D representation of the tongue required manual annotation of the \ac{mri} data, which makes it less feasible for large collections of data.
Work exists that aims at facilitating tongue shape extraction from \ac{mri} data.
However, such approaches are often limited because they are restricted to 2D \citep{Peng2010, Eryildirim2011, Raeesy2013}, produce only a low-level volume segmentation \citep{Lee2013}, or require an anatomical expert to provide the tongue templates \citep{Harandi2014Taylor}.

\subsection{Our contribution}

In this paper, we present a significant extension of our previous work \citep{Hewer2014}.
Originally, we combined an image segmentation method and a template matching approach to extract tongue meshes from \ac{mri} data in a minimally supervised way.
The new features of our framework can be summarized as follows:
we use an image denoising method to deal with possibly corrupt data.
Moreover, we modify the template matching approach to better handle volumetric point clouds.
Additionally, the user can provide landmarks to guide the template matching process.
Furthermore, we integrate a strategy for restoring any tongue surface information that might be missing due to contact between the hard palate and tongue.
This improvement increases the number of tongue shape configurations we can register.
Additionally, the framework is augmented by use of a bootstrapping strategy, which refines the quality of the obtained shape meshes.
Finally, it can now be used to derive a multilinear statistical model that captures almost the entire complex 3D surface geometry of the tongue and allows the anatomy and pose related variations to be modified separately.

We use our new framework to register speech related tongue shapes of 11 speakers and examine the obtained model with respect to its compactness, generalization, and specificity properties.
In the case of the specificity analysis, we investigate those parts of the tongue surface mesh that play an important role during human articulation.
The results of our experiments motivate us to choose a model with 5 \ac{dof} for the anatomy and 4 for the speech related tongue pose.
Moreover, we successfully use the obtained model for tracking motion capture data of the tongue.

The remainder of the paper is organized as follows:
in the next section, we start by describing how surface information of the vocal tract can be extracted from a given 3D \ac{mri} scan by denoising it and applying an image segmentation approach.
We proceed by discussing the modified template matching approach in \autoref{sec:template-matching} and also present the used templates of our approach.
\autoref{sec:shape-estimation} is dedicated to describing how we estimate a tongue mesh from the surface information by using the template fitting.
In this part, we present the bootstrapping strategy used, and our approach to restore missing tongue surface information that is caused by contact between tongue and hard palate.
Next, we turn to the multilinear tongue model in \autoref{sec:multilinear-tongue-model}.
In this section, we outline how the acquired mesh collection can be aligned to only contain speech and anatomy related tongue shape variations, and how the model is derived.
We then turn to the evaluation of our approach in \autoref{sec:evaluation}, where we apply it to \ac{mri} scans of two datasets.
Afterwards, we investigate the validity of our obtained mesh collection by means of a preference test in \autoref{sec:evaluation-mesh-extraction}.
Furthermore, we conduct experiments to evaluate the compactness, generalization, and specificity properties of the acquired model in \autoref{sec:model-analysis}.
In \autoref{sec:tracking}, we use the model for tracking speech related motion capture data of a new speaker.
Finally, we conclude in \autoref{sec:conclusion} and outline possible future work.

\section{Extracting surface information from MRI}
\label{sec:surface-extraction}

As a first step, we want to extract a point cloud \(Q := \{(\vec{q}_i, \vec{n}_i)\}\) from an \ac{mri} scan that contains the surface points \(\vec{q}_i\) and the associated normals \(\vec{n}_i\) of the major articulators and related tissue.
We use a purely geometric representation of this surface information because it is easy to combine two point clouds into a single one.
This is helpful in situations where we want to restore missing information in a point cloud \(Q\) that is present in another cloud \(Q^*\).

As we are using image processing methods, we briefly describe how we treat a volumetric \ac{mri} scan as a 3D image.
We may represent an \ac{mri} scan as a function

\begin{equation}
  \label{eq:scan-function}
  s : \Omega \to \left[s_\text{min}, s_\text{max}\right]
\end{equation}
where \(s_\text{min}\) and \(s_\text{max}\) are real values.
Here, \(\Omega \subset \mathbb{R}^3\) is a discrete rectangular domain consisting of the sample positions where the scanner took the measurements.
These coordinates are arranged on a regular grid with grid spacings \(h_x, h_y,\) and \(h_z\).
We say that \(s(\vec{q})\) represents the measured \ac{nmr}%
\footnote{correlated with hydrogen molecule density, \ie, high for soft tissue, low for bone and air}
at sample position \(\vec{q} \in \Omega\).
This scan can be interpreted as a gray-scale 3D image
\begin{equation}
  \label{eq:image-function}
  f : \Omega \to [0, 255]
\end{equation}
by applying a quantization operator to the \ac{nmr} values that scales them to a standard (\SI{8}{\bit}) gray-scale.
We decided to use a standard visualization where bright and dark indicate a high and low \ac{nmr}, respectively.

\begin{figure}
  \begin{subfigure}{.5\linewidth}
    \includegraphics[scale=0.9]{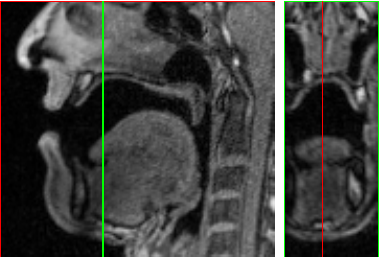}
    \caption{Raw \ac{mri} images}
    \label{fig:smoothing-raw}
  \end{subfigure}
  \begin{subfigure}{.5\linewidth}
    \hfill
    \includegraphics[scale=0.9]{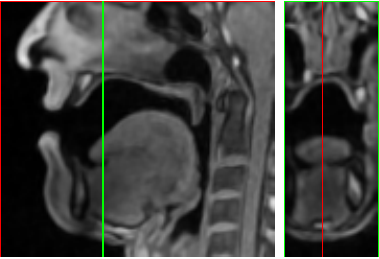}
    \caption{Smoothed \ac{mri} images}
    \label{fig:smoothing-smoothed}
  \end{subfigure}
  \caption{Raw \ac{mri} scan (\subref{fig:smoothing-raw}) and smoothed version (\subref{fig:smoothing-smoothed}).
    The left image of each pair shows a sagittal slice, the other one a coronal slice.
  }
  \label{fig:smoothed-scan}
\end{figure}

\autoref{fig:smoothed-scan} shows two typical visualizations of such a representation:
a sagittal slice and a coronal one showing an \((x, y)\)-plane and a \((y, z)\)-plane of the scan image, respectively.
As in general the original scan \ac{fov} contains much more information than just the vocal tract, we usually crop it to a selected \ac{roi}.
This reduces the memory requirements and the processing time of our framework.

By inspecting the scan, we observe that the data is degraded due to measurement noise.
As a remedy, we apply a 3D variant of edge-enhancing diffusion \citep{Weickert1998} to the image.
An example result of the approach can be inspected in \autoref{fig:smoothing-smoothed}.
We see that the noise was removed and structural information like edges were preserved and enhanced.

\begin{figure}
  \centering
  \begin{subfigure}[t]{0.3\linewidth}
    \includegraphics[width=\linewidth]{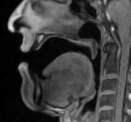}
    \caption{One sagittal slice}
    \label{fig:example-mri}
  \end{subfigure}
  \hfill
  \begin{subfigure}[t]{0.3\linewidth}
    \includegraphics[width=\linewidth]{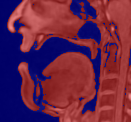}
    \caption{Slice segmentation}
    \label{fig:example-segmentation}
  \end{subfigure}
  \hfill
  \begin{subfigure}[t]{0.3\linewidth}
    \includegraphics[width=\linewidth]{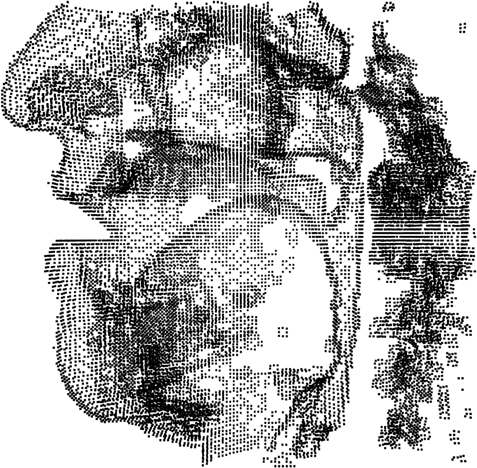}
    \caption{Resulting point cloud}
    \label{fig:example-pointcloud}
  \end{subfigure}
  \caption{Extracting surface information from an \ac{mri} scan: one sagittal slice (\subref{fig:example-mri}), its segmentation (\subref{fig:example-segmentation}), and a rendering of the resulting point cloud (\subref{fig:example-pointcloud}). For visibility reasons, the cloud was clipped and decimated. }
  \label{fig:example-segmentation-and-pointcloud}
\end{figure}

We now want to extract a point cloud \(Q\) of the desired surface information from the denoised \ac{mri} scan.
First, we detect the spatial support of the region whose surface information we want to derive.
That is, we want to find a partition
\begin{equation}
  \label{eq:partition}
  \Omega = \Omega_O \cup \Omega_B
\end{equation}
such that \(\Omega_O\) contains the region of the major articulators and related tissue and \(\Omega_B = \Omega \setminus \Omega_O\) everything else.
By inspecting the denoised data, we notice that tissue can be distinguished from non-tissue, such as air and bone, by using color information.
This observation motivates the use of image segmentation methods that make use of such a feature.
In our case, we decided to use the method introduced by \citet{Otsu1975} to perform this task as it is fully automatic.
An example segmentation can be seen in \autoref{fig:example-segmentation}.

As we are interested in the shape information of the surface, we proceed by extracting the surface points of the tissue from the obtained partition.
We call \(\vec{q}_i \in \Omega_O \) a surface point if at least one of its neighbors is part of \(\Omega_B\).
Additionally, we use the partition to estimate normal information at the extracted surface points.

The obtained surface points and associated normals are then assembled in a point cloud.
An example of such a point cloud can be inspected in \autoref{fig:example-pointcloud}.

\section{Template matching}
\label{sec:template-matching}

Next, we want to estimate the surface of the desired articulator from such a point cloud \(Q\).
Here, we use a polygon mesh \(M := (V, F)\) as the surface representation.
The set \(V := \{\vec{v}_i\}\) with \(\vec{v}_i \in \mathbb{R}^3\) is called the vertex set of the mesh.
The other set, \(F\), is the face set of our mesh.

We observe that a point cloud \(Q\) is a loose collection of points.
In general, this collection contains more information than the desired articulator and there might be holes in the point cloud with missing data.
However, a subset of \(Q\) implicitly represents the surface of the desired articulator.

In order to identify this subset and estimate the articulator shape from it, we can apply a template fitting technique.

Given a template mesh \(M = (V, F)\) that resembles the desired articulator and a point cloud \(Q\), it finds a set \(A := \{A_i\}\) where \(A_i: \mathbb{R}^3 \to \mathbb{R}^3\) is a rigid body motion for the vertex \(\vec{v}_i \in V\), such that the deformed mesh \(M^* = (V^*, F)\) with \(V^* := \{A_i(\vec{v}_i)\}\) is near the point cloud data \(Q\).

The template matching finds this set \(A\) of deformations by minimizing the energy
\begin{equation}
  \label{eq:energy-template-matching}
  E_{\text{Def}}(A) = \alpha\ E_\text{data}(A) + \beta\ E_\text{smooth}(A) +  \gamma\ E_\text{landmark}(A)
\end{equation}
The data term
\begin{equation}
    E_\text{data}(A) := \frac{1}{|V^L|} \sum_{\vec{v}_i \in V^L} \Big\| A_i(\vec{v}_i) - \operatorname*{arg\ min}_{\vec{p}_j \in Q}  \| A_i(\vec{v}_i) - \vec{p}_j\| \Big\|^2
\end{equation}
measures the distance between the deformed vertices \(A_i(\vec{v}_i)\) and their nearest neighbors \(\vec{p}_j\) in the point cloud \(Q\).
Thus, it is minimized if applying \(A\) to the mesh moves it towards some points in the point cloud.
In this term, \(V^L\) refers to the set of vertices that are not landmarks.
The smoothness term
\begin{equation}
    E_\text{smooth}(A) := \frac{1}{|V|} \sum_{v_i \in V} \Bigg( \sum_{v_j \in \mathcal{N}(v_i)} \big\|A_i - A_j\big\|^2 \Bigg)
\end{equation}
evaluates the differences between the rigid body motion \(A_i\) at vertex \(\vec{v}_i\) and the motions \(A_j\) in its neighborhood \(\mathcal{N}(\vec{v}_i)\).
This means that it penalizes deformations that alter the original shape of the template.
Finally, the landmark term
\begin{equation}
    E_\text{landmark}(A) := \frac{1}{|L|} \sum_{(\vec{v}_i, \vec{p}_i) \in L} \big\| A_i(\vec{v}_i) - \vec{p}_i \big\|^2
\end{equation}
produces energy in proportion of how many correspondences between deformed landmark vertices \(A_i(\vec{v}_i)\) and user-provided target points \(\vec{p}_i\) of the landmark set \( L := \{ (\vec{v}_i, \vec{p}_i)\}\) are violated by the deformation.
We remark that the used target points are not required to be part of the generated point cloud.
As a convention, we always set the weight \(\alpha\) to 1 in order to interpret the other parts of the energy in terms of the data nearness assumption:
for example, using a value of \(\beta = 10\) means that the smoothness term is ten times more important than the data term. 

As the energy in \eqref{eq:energy-template-matching} is not differentiable due to the data term, it is usually optimized by minimizing a series of energies \(E^t_\text{Def}(A^t)\) where \(t \in [1, t_\text{max}]\).
Each energy uses adapted weights \(\beta^t\) and \(\gamma^t\):
\begin{align}
  \beta^t =& \beta - ( t - 1 ) \frac{\beta - \beta_\text{min}}{t_{\text{max}} - 1} \\
  \gamma^t =& \gamma - ( t - 1 ) \frac{\gamma - \gamma_\text{min}}{t_{\text{max}} - 1}
\end{align}
where \(\beta_\text{min}\) and \(\gamma_\text{min}\) are set by the user.

\begin{figure}
  \centering
  \begin{subfigure}{.3\linewidth}
    \centering
    \includegraphics[width=0.8\linewidth]{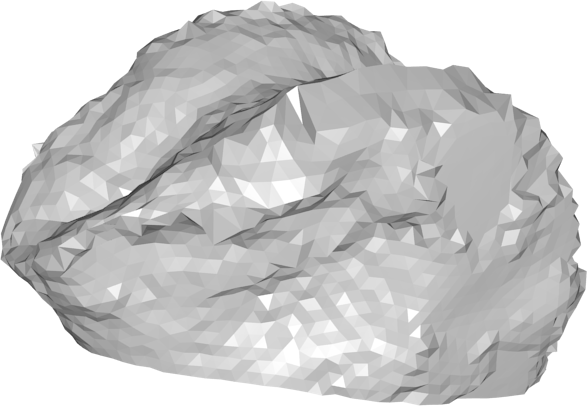}\\[10pt]
    \includegraphics[width=0.8\linewidth]{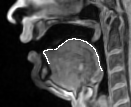}
    \caption{\(\beta_\text{min} = 0.1\)}
    \label{fig:beta-effect-0.1}
  \end{subfigure}
  \begin{subfigure}{.3\linewidth}
    \centering
    \includegraphics[width=.8\linewidth]{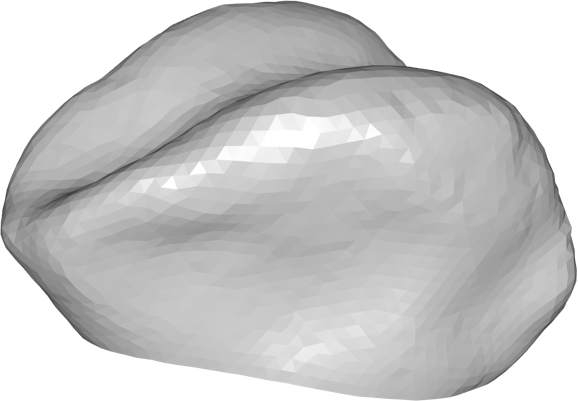}\\[10pt]
    \includegraphics[width=.8\linewidth]{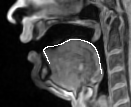}
    \caption{\(\beta_\text{min} = 5\)}
    \label{fig:beta-effect-5}
  \end{subfigure}
  \begin{subfigure}{.3\linewidth}
    \centering
    \includegraphics[width=.8\linewidth]{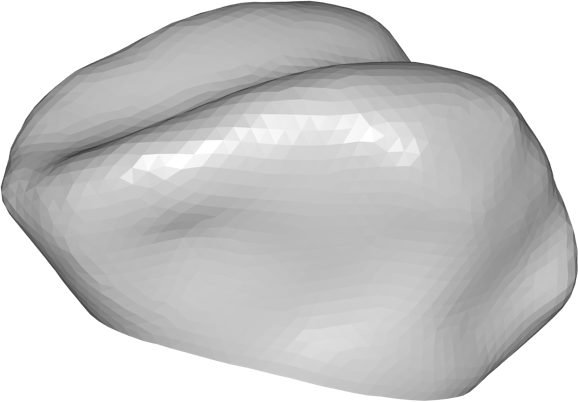}\\[10pt]
    \includegraphics[width=.8\linewidth]{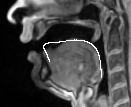}
    \caption{\(\beta_\text{min} = 100\)}
    \label{fig:beta-effect-100}
  \end{subfigure}
  \caption{Effect of \(\beta_\text{min}\) on the resulting mesh.
    A low value (\subref{fig:beta-effect-0.1}) leads to an overfitting of the data and a very noisy mesh, whereas a high value causes underfitting and produces a very smooth result (\subref{fig:beta-effect-100}).
    Choosing an appropriate value provides a good compromise between data nearness and mesh quality (\subref{fig:beta-effect-5}).}
  \label{fig:beta-effect}
\end{figure}

These parameters have to be carefully selected as they influence the last energy that is optimized.
A value of \(\beta_\text{min}\) that is too high forces the approach to preserve the shape of the original template, which leads to an underfitting.
Setting the value too small, on the other hand, causes an overfitting that produces many local shape artifacts on the resulting mesh.
\autoref{fig:beta-effect} shows results for different values of \(\beta_\text{min}\).

\begin{figure}
  \centering
  \begin{subfigure}[t]{.3\linewidth}
    \centering
    \includegraphics[width=0.8\linewidth]{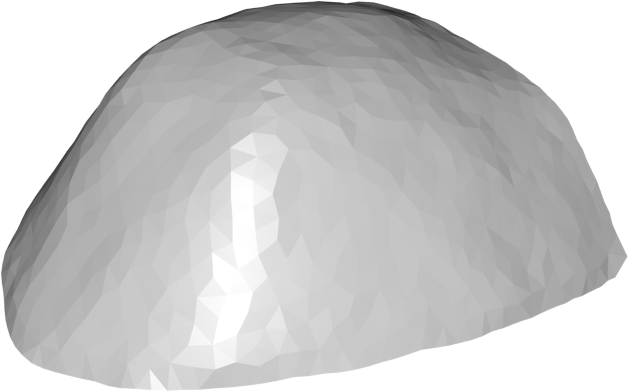}\\[10pt]
    \includegraphics[width=0.8\linewidth]{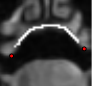}
    \caption{\(\gamma_\text{min} = 0\)}
    \label{fig:gamma-effect-0}
  \end{subfigure}
  \begin{subfigure}[t]{.3\linewidth}
    \centering
    \includegraphics[width=.8\linewidth]{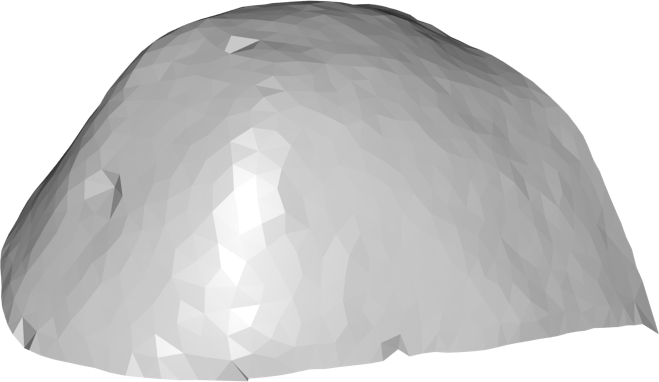}\\[10pt]
    \includegraphics[width=.8\linewidth]{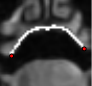}
    \caption{\(\gamma_\text{min} = 10\)}
    \label{fig:gamma-effect-10}
  \end{subfigure}
  \begin{subfigure}[t]{.3\linewidth}
    \centering
    \includegraphics[width=.8\linewidth]{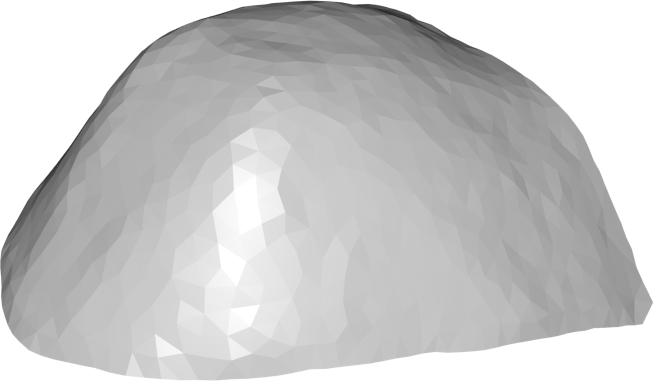}\\[10pt]
    \includegraphics[width=.8\linewidth]{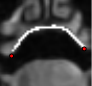}
    \caption{\(\gamma_\text{min} = 10\) and post smoothing}
    \label{fig:gamma-effect-10-smooth}
  \end{subfigure}
  \caption{Effect of \(\gamma_\text{min}\) on the resulting mesh.
    Setting it to \(0\) prevents the template matching from reaching the provided landmarks shown as red dots (\subref{fig:gamma-effect-0}).
    Using the value \(10\) aligns the template to the wanted positions, but leads to spike-like artifacts (\subref{fig:gamma-effect-10}).
    Applying a smoothing afterwards removes these spikes while keeping the template close to the landmarks (\subref{fig:gamma-effect-10-smooth}).
  }
  \label{fig:gamma-effect}
\end{figure}

A similar statement holds true for \(\gamma_\text{min}\):
using too small a value could move the template away from the desired landmark locations during the optimization.
A value that is too high might overfit the landmark positions, which could cause problems if landmarks are wrongly placed, and lead to spike-like artifacts.
In our approach, we mitigate such effects of wrongly placed landmarks by applying a smoothing operation after the template matching:
we replace the measured rigid body motion \(A_i\) at the corresponding vertex with the average of the rigid body motions of vertices that are connected via an edge to this vertex.
In \autoref{fig:gamma-effect}, the effect of \(\gamma_\text{min}\) on the mesh can be inspected.
Furthermore, we show how the smoothing operation improves the result.
It is important to avoid noise or artifacts on the mesh because we want to prevent our tongue model from modeling this noise.

Originally, we used a standard heuristic \citep{Allen2003, Li2009} to distinguish valid data observations from invalid ones in the optimization of \(E_\text{data}\).
In particular, we say that \(\vec{q}\) is a valid data point candidate for a deformed vertex \(A_i(\vec{v}_i)\) if the Euclidean distance between both is not too large and if their normals do not differ too much from each other.

In this paper, we have modified this nearest neighbor heuristic somewhat:
we now collect all valid data point candidates within a fixed radius and then select the best candidate that lies below the current mesh surface.
If no such candidate exists below the surface, we will select the best one above it.
This modification is intended to prevent the template mesh from getting stuck at unrelated points in the volumetric cloud during the optimization.

In our framework, we use two templates:
one for the tongue and one for the hard palate.
Both templates were extracted from \ac{mri} data by means of medical imaging software \citep{Rosset2004}.
Afterwards, we made the templates symmetric to remove this particular bias towards the original speaker by mirroring the respective mesh at a selected center plane.

The palate template consists of \num{994} vertices and \num{1828} faces with an average edge length of \SI{1.4}{\mm}.
The tongue template contains \num{3100} vertices and \num{6102} faces with an average edge length of \SI{1.8}{\mm}.
In our case, the tongue template does not contain the sublingual part.
This means that the part below the line from the jaw to the epiglottis is missing, as well as the part below the tongue tip that is negligible for speech production.

\begin{figure}
  \begin{subfigure}[t]{.5\linewidth}
    \includegraphics{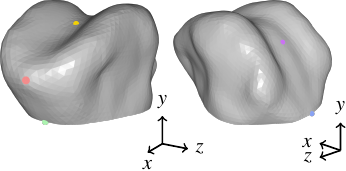}
  \end{subfigure}
  \begin{subfigure}[t]{.5\linewidth}
    \hfill
    \includegraphics{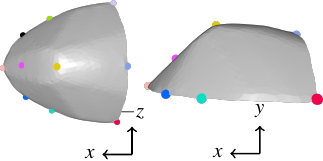}
  \end{subfigure}
  \caption{Used templates with landmarks of the tongue (left) and hard palate (right).}
  \label{fig:templates}
\end{figure}

Both templates can be inspected in \autoref{fig:templates} together with the landmarks used.

\section{Tongue and palate shape estimation}
\label{sec:shape-estimation}

We first estimate the palate shape for each \ac{mri} scan.
We select a scan for each speaker where the hard palate is clearly visible and perform template matching.
In general, using a single template might produce sub-optimal results in some matching cases.
In order to improve the results, we set up an iterative bootstrapping approach.
In each iteration, we first compute a \ac{pca} model of the palate \citep{Hewer2015} by using the results of the previous iteration.
The palate model is derived in a way similar to the multilinear model in \autoref{sec:multilinear-tongue-model}.
This model is then fitted to each point cloud and the results are afterwards used as the initialization for the template matching.

\begin{figure}
  \begin{subfigure}{.5\linewidth}
    \includegraphics{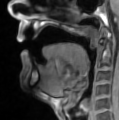}
    \includegraphics{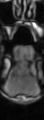}
    \caption{[a] scan with visible palate contour}
    \label{fig:no-palatal-contact-nosurface}
  \end{subfigure}
  \begin{subfigure}{.5\linewidth}
    \hfill
    \includegraphics{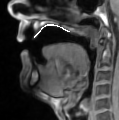}
    \includegraphics{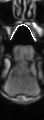}
    \caption{[a] scan with registered palate mesh}
    \label{fig:no-palatal-contact-surface}
  \end{subfigure}
  \begin{subfigure}{.5\linewidth}
    \includegraphics{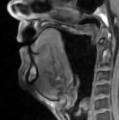}
    \includegraphics{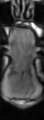}
    \caption{[i] scan with palatal contact}
    \label{fig:palatal-contact-nosurface}
  \end{subfigure}
  \begin{subfigure}{.5\linewidth}
    \hfill
    \includegraphics{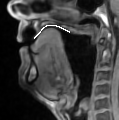}
    \includegraphics{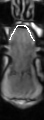}
    \caption{[i] scan with hard palate surface restored}
    \label{fig:palatal-contact-surface}
  \end{subfigure}
  \caption{Palate reconstruction:
    scan of speaker pronuncing [a] (\subref{fig:no-palatal-contact-nosurface}), and registered hard palate surface (\subref{fig:no-palatal-contact-surface}).
    Palatal contact during the pronunciation of [i] (\subref{fig:palatal-contact-nosurface}), and result of restoring hard palate surface information (\subref{fig:palatal-contact-surface}).
    It should be observed that in (\subref{fig:palatal-contact-nosurface}), the hard palate appears to merge with the tongue, which presents a potential pitfall for manual annotation.
    The palate reconstruction, on the other hand, helps to identify the true palate contour.
  }
  \label{fig:palate-reconstruction}
\end{figure}

After we have acquired the hard palate mesh for each speaker, we align this mesh to each scan of that speaker.
This procedure serves the purpose of restoring tongue surface information that is missing due to contacts between tongue and palate as shown in \autoref{fig:palate-reconstruction}.

In this context, we have to address the issue that the corresponding speaker might have moved between the scans.
Fortunately, as the hard palate can only undergo rigid body transformations, we only have to estimate this type of motion.
However, as the palate surface information might be partly missing, we fall back to \ac{nmr} information for this task.
To this end, we define the \ac{nmr} profile set \(E(M, f) \subset \mathbb{R}^\ell\) of a mesh \(M\) in a scan \(f\).
A profile \(\vec{e}^i (M, f) \in E(M, f)\) is a vector such that its entries are given by
\begin{equation}
  \label{eq:color-profile}
  \vec{e}_j^i(M, f) = f( \vec{v}_i + j\ d\ \vec{n}_i)
\end{equation}
where \(\vec{v}_i\) is a mesh vertex, \(\vec{n}_i\) its corresponding normal, and \(d\) the chosen sampling distance.
We start above the palate surface in order to avoid taking samples in the possible contact area between tongue and palate.
We can estimate the rigid body motion \(A\) for aligning a palate mesh \(M\) obtained from a scan \(f\) to a scan \(g\) by maximizing the energy:
\begin{equation}
  \label{eq:palate-alignment}
  E_\text{palate} (A) = \sum_{i \in J(V)} \operatorname{NCC} \left(\vec{e}^i(M, f), \vec{e}^i(A(M), g)\right)
\end{equation}
where \(J(V)\) is the index set of the vertex set \(V\), \(\operatorname{NCC}\) the normalized cross-correlation between its operands, and \(A(M)\) the transformed mesh.
We decided to use the \ac{ncc} as a similarity measure because it is known to be robust against noise and brightness differences.
Furthermore, the \ac{ncc} between \ac{nmr} profiles was already successfully used in a nearest neighbor heuristic for template matching \citep{Harandi2014Taylor}.
Results of this alignment approach can be seen in \autoref{fig:palate-reconstruction}.
In this figure, we also present the result for a scan without contact between tongue and palate, to show the actual contour of the hard palate of the speaker, which is not visible in the other scan.

We now inject this aligned palate mesh information into the point cloud of the corresponding scan in order to restore missing tongue surface information by using the palate surface as a replacement.
Additionally, we use the aligned mesh as a boundary to remove points in the point cloud above the palate that are unrelated to the tongue.
Finally, we use template matching to extract the tongue shape from the corresponding modified point cloud.
As in the palate case, we use a bootstrapping strategy to refine the results.
\begin{figure}
  \begin{subfigure}{.5\linewidth}
    \includegraphics{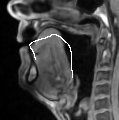}
    \includegraphics{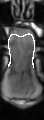}
    \caption{Initial template}
    \label{fig:tongue-bootstrapping-initial}
  \end{subfigure}
  \begin{subfigure}{.5\linewidth}
    \hfill
    \includegraphics{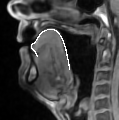}
    \includegraphics{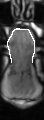}
    \caption{Bootstrapping result}
    \label{fig:tongue-bootstrapping-improved}
  \end{subfigure}
  \caption{Effect of the bootstrapping strategy.
    Initial template matching has trouble correctly registering the tongue shape (\subref{fig:tongue-bootstrapping-initial}).
    Bootstrapping improves the result (\subref{fig:tongue-bootstrapping-improved}).
  }
  \label{fig:tongue-bootstrapping}
\end{figure}
This time, we use a multilinear model in each iteration as a statistical prior that is described in the next section.
The benefits of this bootstrapping operation can be seen in \autoref{fig:tongue-bootstrapping}.

\section{Multilinear tongue model}
\label{sec:multilinear-tongue-model}

Having obtained a collection of tongue meshes, we now want to derive a function
\begin{equation}
  \label{eq:multilinear-model}
  M : S \times P \to \mathbb{M}
\end{equation}
where \(\mathbb{M}\) is a set of meshes.
The set \(S \subseteq \mathbb{R}^{\widetilde{m}}\) consists of coordinates \(\vec{s}\) that describe a speaker's anatomical tongue shape.
The set \(P \subseteq \mathbb{R}^{\widetilde{n}}\) contains coordinates \(\vec{p}\) that determine the shape for a specific speech related tongue pose.
We call \(S\) the speaker subspace and \(P\) the pose subspace of the model.
Meshes \(M \in \mathbb{M}\) should have the same face set as our tongue template mesh.
Their vertex sets \(V(\vec{s}, \vec{p})\), however, may differ from the original template with respect to their vertex positions.

\subsection{Preparing the training mesh collection}

Deriving the function in \eqref{eq:multilinear-model} implies we want to analyze only the anatomical and speech related variations in our mesh collection, which means we have to remove all other variations present.
The Procrustes alignment technique \citep{Dryden1998} is a method suitable for this task as it may be used to remove any translational and rotational differences among the meshes in the collection.
However, applying this technique directly to the acquired tongue meshes might eliminate critical information related to, \eg, the speech related tongue pose.
This is, for example, due to the fact that the tongue also undergoes translational and rotational motions because it is connected to the lower jaw.

As a remedy, we apply the Procrustes alignment to the hard palate meshes we obtained earlier to remove translational and rotational differences between the speakers that are unrelated to the tongue motion.
The results are afterwards used as a reference to align the tongue meshes.
To this end, we use a speaker's palate mesh that was earlier aligned to the corresponding scan.
We then estimate the rigid transformation that maps this aligned palate mesh to its Procrustes variant and apply the same motion to the corresponding tongue mesh.
By doing so, we remove any translational and rotational differences related to head motion or position differences without eliminating any speech or anatomy-specific information.

Finally, we have to ensure that for each speaker the meshes for all selected poses are available.
This might be needed because some \ac{mri} scans may be corrupt or even missing in the used dataset.
In these cases, we reconstruct a missing pose shape of a speaker by averaging available data:
first, we compute the average shape of all meshes that are present for the speaker.
Afterwards, we compute the mean shape of all meshes that are available for this specific pose from the other speakers.
Finally, both meshes are averaged again.
In the literature, there are more sophisticated methods to restore missing information, such as HALRTC \citep{Liu2013}.
In our case, however, this averaging approach was sufficient.

\subsection{Model construction}

In order to derive our desired function in \eqref{eq:multilinear-model}, we need to analyze the anatomical and speech related variations separately.
In previous work \citep{Harshman1977, Hoole2000, Hoole2003, Ananthakrishnan2010, Valdes2012, Vargas2012, Zheng2003}, the PARAFAC method \citep{Harshman1970}, also known as CANDECOMP, has often been used to perform this analysis.
This method decomposes a tensor into a sum of \(r\) rank-\(1\) tensors where \(r\) is provided by the user.
Therefore, this technique can be regarded as an extension of the singular value decomposition to tensors.
However, there are reports in the literature of certain issues with this method:
\citet{Hoole2000} found that it might be difficult to find reliable solutions;
\citet{Vargas2012} pointed out that the PARAFAC decomposition requires numerous components to describe the observed data in a satisfactory way, which limits its usefulness as a dimensionality reduction method;
moreover, \citet{deSilva2008} discovered that the associated standard approximation problem is mathematically ill-posed, which can lead to the problem of diverging components in a numerical setting.

Another suitable method is the Tucker decomposition \citep{Tucker1966}, which is sometimes also called \ac{hosvd}.
This method computes the orthonormal spaces of a tensor associated with its modes.
It may be regarded as a more flexible variant of the PARAFAC method \citep{Kiers1991} and has previously been used to analyze 2D tongue shape data \citep{Valdes2012}.
To avoid the issues with PARAFAC, we decided to use the Tucker decomposition to analyze our data.
We follow the approach of \citet{Bolkart2015} who used it to analyze the variations of human faces in different expressions.
To this end, we first turn our tongue meshes into feature vectors by serializing the vertex sets \(V\) into vectors \(\vec{f}_i\).
Then, we compute the mean \(\mu\), and center the vectors.
Afterwards, we organize those centered vectors in a tensor \(A \in \mathbb{R}^{m \times n \times k}\).
Here, we refer to the first mode of the tensor as the \emph{speaker} mode where \(m\) represents the number of speakers, to the second mode as \emph{pose} mode with \(n\) being the number of different tongue poses, and to the third mode as the \emph{vertex} mode with \(k\) representing the dimension of the vectors \(\vec{f}_i\).

The \ac{hosvd} makes use of the fact that \(A\) can be decomposed as follows:
\begin{equation}
  \label{eq:tensor-decomposition}
  A = C \times_1 U_1 \times_2 U_2
\end{equation}
In our case, the row vectors of \(U_1 \in \mathbb{R}^{m \times m}\) are coordinates in our speaker space \(S\) that determine the anatomical shape for each of the original speakers.
A similar observation applies to \(U_2 \in \mathbb{R}^{n \times n}\) where the row vectors are coordinates in the pose space \(P\).
The tensor \(C \in \mathbb{R}^{m\times n \times k}\) is the core tensor of the decomposition that acts as a link between \(S\) and \(P\).
The operation \(C \times_n U\) is called the \(n\)-th mode multiplication of the tensor \(C\) with the matrix \(U\).

The core tensor is the multilinear model we can use to create our function in \eqref{eq:multilinear-model}:
essentially, given \(\vec{s} \in S\) and \(\vec{p} \in P\), we can use \(C\) to generate serialized vertex sets that represent the generated shape as follows:
\begin{equation}
  \label{eq:generation}
  v(\vec{s}, \vec{p}) = \mu + C \times_1 \vec{s} \times_2 \vec{p}
\end{equation}
By letting \(V(\vec{s}, \vec{p})\) be the vertex set reconstructed from \(v(\vec{s}, \vec{p})\), we can finally define our function as:
\begin{equation}
  \label{eq:final-function}
  M(\vec{s}, \vec{p}) := \left(V(\vec{s}, \vec{p}), F\right)
\end{equation}
where \(F\) is the face set of our original template.
We remark that the dimensionality of the speaker and pose subspaces can be truncated to remove shape variations that may be considered negligible or related to noise.
This means that our subspaces have dimensionalities \(\widetilde{m} \leq m\) and \(\widetilde{n} \leq n\).

\subsection{Model fitting}

We can use this derived model to register data, for example a point cloud \(Q\).
This time, we want to optimize for the parameters \(\vec{s} \in S\) and \(\vec{p} \in P\) that best describe the speaker anatomy and tongue pose that is represented in the data.
To this end, we minimize the following energy:
\begin{align}
  E_{\text{Fit}}(\vec{s}, \vec{p}) = \alpha\ E_\text{data}(\vec{s}, \vec{p}) + \gamma\ E_\text{landmark}(\vec{s}, \vec{p})
  \label{eq:fit-energy}
\end{align}
where the data and landmark terms are equivalent in their modeling idea to their counterparts in the template matching case.
Furthermore, we use the same nearest neighbor heuristic and optimization approach as in the template matching.
This time, the weights for both terms remain constant during the optimization of the energy series.
However, if the corresponding neighbor for each vertex is known, they can be set directly and only one energy has to be minimized in that case.
This is the case, for example, if our target is a tongue mesh of the same size, such that a one-to-one correspondence between model mesh and target mesh exists.

It is common to limit the admissible values for \(\vec{s}\) and \(\vec{p}\) to avoid highly unlikely shapes.
In particular, we limit each entry of \(\vec{s}\) and \(\vec{p}\) individually to an interval
\begin{equation}
  \label{eq:value-limitation}
  \left[m_i - h\ \sigma_i, m_i + h\ \sigma_i\right]
\end{equation}
where \(\sigma_i\) is the standard deviation of the corresponding variation in the used mesh collection, and \(m_i\) is the corresponding entry of the mean coordinate in the respective subspace.
Finally, \(h \in \mathbb{R}^+\) is a scale factor.

The above energy can also be used to fit a \ac{pca} model:
in this case, the energy depends only on one parameter.

\section{Deriving a tongue model}
\label{sec:evaluation}

Our next goal is to apply the described framework to \ac{mri} data and obtain a tongue model.
We do not undertake a quantitative evaluation of the tongue meshes extracted from the \ac{mri} scans, a decision that is necessitated by the fact that we are working with real-world data, and thus have no ground truth reference available.
There exists the possibility of manually annotating the \ac{mri} scans to create a reference solution, but this procedure is very time consuming and expensive if the number of scans in the dataset is very large.
Moreover, such hand-labeling is error prone and the anatomical expert(s) involved may introduce a subjective bias.
Instead, we chose to perform a qualitative analysis:
we inspected the results manually as a post-hoc analysis in order to decide whether they are acceptable.
In particular, we projected the registered tongue meshes onto to the corresponding scans and verified that the mesh surface was close to the true tongue contour.
Videos showing such projections slice by slice can be found in the supplementary material.
The corresponding file names start with ``01MRIM\_projection''.

\subsection{Background information about data used}

\begin{figure}
  \centering
  \begin{tabulary}{\linewidth}{CCC}
    10MRIF & 11MRIM & 14MRIF \\
    \includegraphics[height=0.18\textwidth]{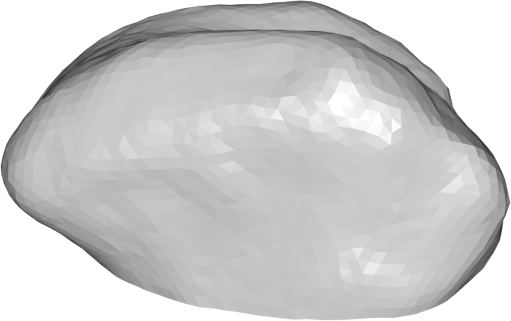} & 
    \includegraphics[height=0.18\textwidth]{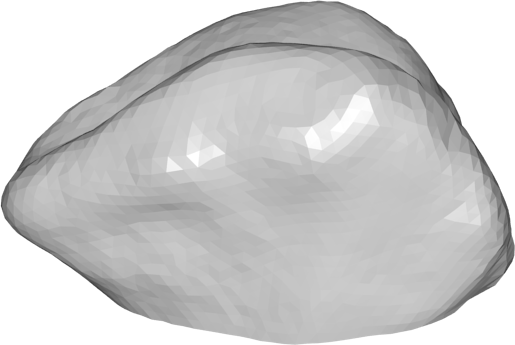} &
    \includegraphics[height=0.18\textwidth]{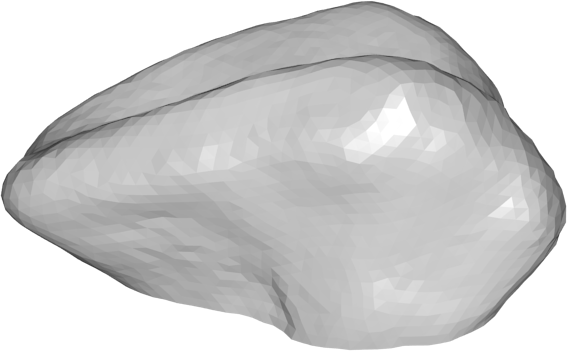} \\
    \includegraphics[height=0.18\textwidth]{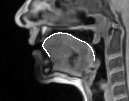} & 
    \includegraphics[height=0.18\textwidth]{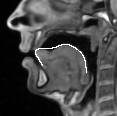} &
    \includegraphics[height=0.18\textwidth]{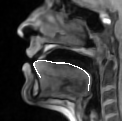} \\
    \includegraphics[height=0.18\textwidth]{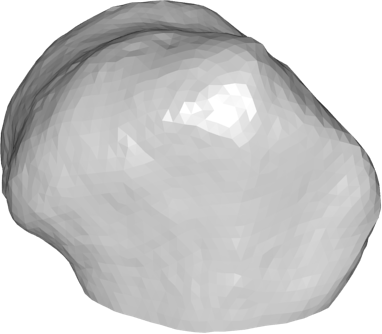} & 
    \includegraphics[height=0.18\textwidth]{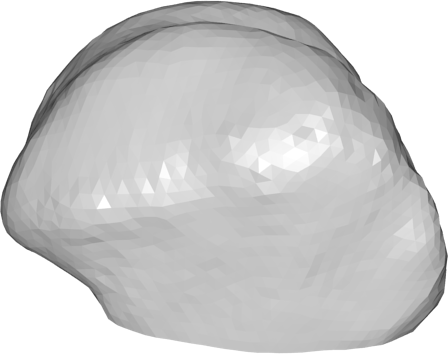} &
    \includegraphics[height=0.18\textwidth]{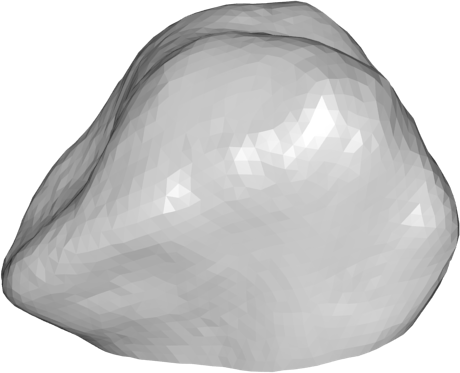} \\
    \includegraphics[height=0.18\textwidth]{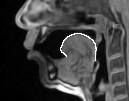} & 
    \includegraphics[height=0.18\textwidth]{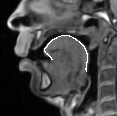} &
    \includegraphics[height=0.18\textwidth]{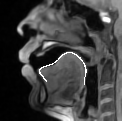} \\
  \end{tabulary}
  \caption{Shape variabilities among the speakers 10MRIF, 11MRIM, and 14MRIF in the dataset used.
    The images show the extracted mesh above a sagittal slice of the corresponding scan for the phones [s] (top) and [u] (bottom).}
  \label{fig:dataset-variability}
\end{figure}

We use two datasets to derive our model:
the dataset of \citet{Baker2011} and the full dataset of the Ultrax project \citep{Ultrax2014}, which provides us with data of 12 speakers in total.

The Ultrax dataset consists of static \ac{mri} scans of 11 adult speakers of British English where 7 are female and 4 are male.
All speakers are phonetically trained and were recorded while sustaining the vocal tract configuration for different phones for around \SI{20}{\s}.
For each speaker, 13 speech related scans are available that correspond to the phone set [i, e, ɛ, a, ɑ, ʌ, ɔ, o, u, ʉ, ə, s, ʃ].%

The Baker dataset was recorded as part of the Ultrax project, but released separately.
It contains 25 scans of one male speaker that are speech related and represent different articulatory configurations.

The data was recorded at the Clinical Research Imaging Centre in Edinburgh using a Siemens Verio 3T scanner;
the scans were acquired with an echo time of \SI{0.93}{\ms} and a repetition time of \SI{2.36}{\ms}.
The individual scans consist of 44 sagittal slices with a thickness of \SI{1.2}{\mm} and a slice size of \num{320 x 320} pixels.
The grid spacings are \(h_x = h_y = \SI{1.1875}{\mm}\) and \(h_z = \SI{1.2}{\mm}\).

For our analysis, we decided to exclude one speaker of the Ultrax dataset that showed a high activity of the soft palate, which caused problems in our framework.
Furthermore, we use the whole phone set that was recorded for the Ultrax data.
However, the Baker dataset is lacking scans for the phones [a, ɔ, ʉ, ə, s, ʃ] where the shape information has to be reconstructed.

In total, we are using the shape information of 11 speakers with 13 different tongue shape configurations.

In \autoref{fig:dataset-variability}, some of these shapes can be seen.
This means that we arrive at a tensor \( A \in \mathbb{R}^{11 \times 13 \times 9300}\) where the dimension of the vertex mode is determined by the vertex count of the tongue template we are using.

It is important to state that a tongue model derived from this data might lack some typical tongue pose related variations because the phonetic coverage of the underlying dataset can be considered incomplete.
Due to these data related constraints, we believe that this model's applicability to some of the application areas we described earlier, may be limited.
However, assembling and using \ac{mri} recordings is a demanding task:
First, it requires access to an \ac{mri} scanner, along with specialized equipment and technical staff experienced in performing such recordings.
Second, appropriate speakers have to be found who are phonetically trained and whose articulation is not impacted by the \ac{mri} scanner.
Moreover, the use and distribution of acquired medical imaging data is governed by strict and extensive data privacy protection:
For example, the raw data cannot normally be published, and using it for research purposes requires the explicit consent of the corresponding speaker.
Under these considerations, we argue that even a limited dataset is a valuable resource and can lead to useful results.

\subsection{Applied settings}

In the following, we describe the settings we applied in our framework to extract the mesh collection from the given data.
In general, the settings were chosen in such a way that the results were visually satisfactory.
This means that we wanted the meshes to be close to the tongue contour that is visible in the \ac{mri} scan.
Moreover, we tried to reduce overfitting artifacts, such as spikes or noise, on the resulting shapes.

In the case of template matching, we used \(\alpha = 1\), \(\beta = 10\), \(\beta_\text{min} = 6\), and \(\gamma = 10\).
Thus, we start with a high weight for the smoothness and landmark terms to drive the template to the correct neighborhood at the beginning of the optimization.
The template matching for the tongue used \(\gamma_\text{min} = 0\) to dampen the effects of falsely placed landmarks.
We used \(\gamma_\text{min} = 10\) for the palate matching to ensure that its extremities were correctly aligned.
For the model fitting that is applied during the bootstrapping, we used \(\alpha = \gamma = 1\).
In the nearest neighbor heuristic, we set the search radius to \SI{4}{\mm} and limited the maximally allowed angle difference between the normals to 60 degrees.
The optimization for the template matching used a series of 40 energies, the one for the model fitting applied a series of 10 energies to find the minimizer.
For the palate alignment, we decided to use sufficiently long profiles with a length of \(\ell = 15\) and a sampling distance of \(d = \SI{1}{\mm}\).

In the bootstrapping strategy, we applied the following iteration amounts:
we used one iteration for the hard palate and 5 iterations for the tongue.
For the scale factor \(h\) in the model fitting, we used \(0.5\) for the tongue and \(1\) for the palate in order to prevent overfitting.

The landmarks needed for the hard palate and the tongue were placed on the \ac{mri} scans by a user who is not an anatomical expert.

\section{Evaluation of the mesh extraction process}
\label{sec:evaluation-mesh-extraction}

We mentioned earlier that no ground truth is available for our dataset.
Before analyzing and using the acquired tongue model, however, we wanted to evaluate the validity of the meshes we extracted from the \ac{mri} data.
To this end, we designed a web-based preference test and elicited the opinion of speech experts.
The goal of this experiment was to investigate whether the experts agreed with our own informal assessment of the acquired meshes.

\subsection{Experiment setup}

We prepared the following data for the experiment:
for each of the 137 scans in our dataset, we created three versions of the same sagittal slice.
One version showed the unannotated slice.
The second version showed the slice with the tongue mesh contour after the initial template matching.
The last version visualized the tongue mesh contour after the final bootstrapping.
Afterwards, we randomly partitioned our scan set into 4 subsets of roughly equal size.
These partitions were then randomly assigned to the participants such that overall, each scan was seen by \numrange{3}{4} participants.

15 speech experts took part in the experiment.
On average, they had 11 years of research experience with speech production data.
Each participant was asked to view all scans of the assigned partition and to select the preferred annotated version of the shown sagittal slices.
During the experiment, the individual methods that produced the results were hidden from the participants.
Moreover, in order to prevent the participants from detecting any pattern in the presentation, the two annotated versions were always displayed in random order.

\subsection{Results}

\begin{figure}
  \centering
  \begin{subfigure}[t]{0.3\linewidth}
  \centering
    \includegraphics[height=0.8\linewidth]{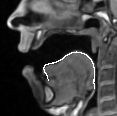}
    \\[.5\baselineskip]
    \includegraphics[height=0.8\linewidth]{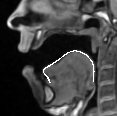}
    \caption{[ɑ] scan of 11MRIM}
  \end{subfigure}
  \hfill
  \begin{subfigure}[t]{0.3\linewidth}
  \centering
    \includegraphics[height=0.8\linewidth]{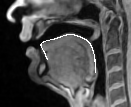}
    \\[.5\baselineskip]
    \includegraphics[height=0.8\linewidth]{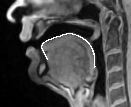}
    \caption{[ʌ] scan of 01MRIM}
  \end{subfigure}
  \hfill
  \begin{subfigure}[t]{0.3\linewidth}
  \centering
    \includegraphics[height=0.8\linewidth]{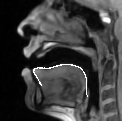}
    \\[.5\baselineskip]
    \includegraphics[height=0.8\linewidth]{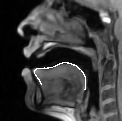}
    \caption{[ə] scan of 14MRIF}
  \end{subfigure}
  \caption{Examples for scans where the participants preferred the initial template matching result (top row) over the bootstrapping one (bottom row).}
  \label{fig:survey-similar-results}
\end{figure}

The evaluation revealed that in \SI{83.85}{\percent} of the cases, the bootstrap result was preferred by the participants.
We proceeded by investigating how these preferences were distributed among the different scans that were shown.

\begin{figure}
  \centering
  \includegraphics{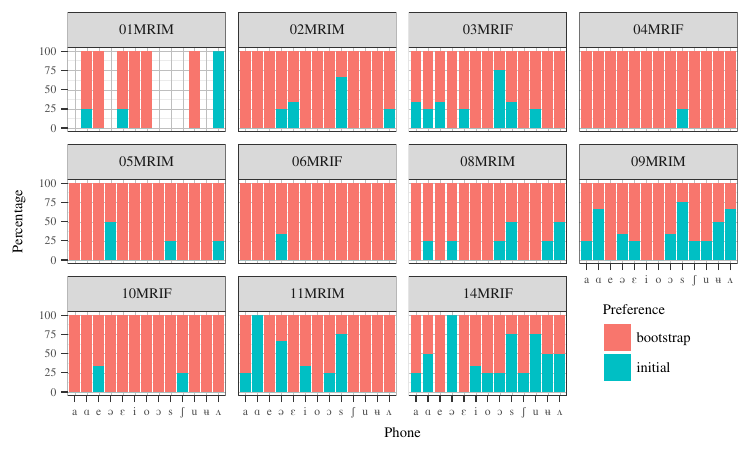}
  \caption{Results of the preference test for each considered scan.
    Note that not all phonemes are available in the data for speaker 01MRIM.
    We grouped the scans by speaker to improve the visualization.
  }
  \label{fig:survey-results}
\end{figure}

A plot summarizing our findings can be seen in \autoref{fig:survey-results}.
We see that for 19 scans, in \SI{50}{\percent} or more cases, the initial template matching was preferred over the bootstrapping result.
For these individual scans, we inspected the displayed slices and discovered that the initial and bootstrapping versions were very similar.
Moreover, the bootstrapping results seemed to slightly underestimate the tongue shape in the \ac{mri} scan in these cases, which might have caused the participants to choose the initial result.
Examples of such cases are shown in \autoref{fig:survey-similar-results}.

\subsection{Discussion}

Overall, we draw two conclusions from the obtained results of the experiment.
On the one hand, the relatively high acceptance rate of the obtained bootstrap results among the consulted experts justifies our decision to derive a tongue model from the corresponding mesh collection.
On the other hand, however, we see also that there is still some room for improvement of our approach, which is conditioned on speaker-specific anatomy.

\section{Model analysis}
\label{sec:model-analysis}

\begin{figure}
  \centering
  \includegraphics{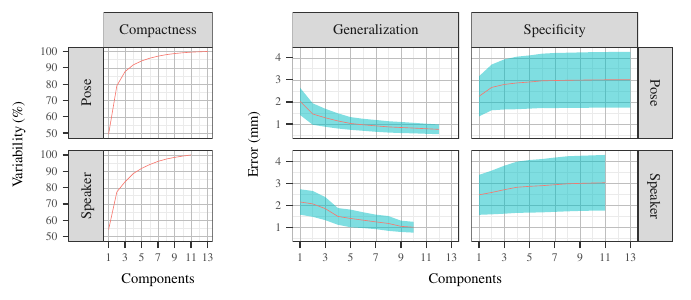}
  \caption{Compactness (left), generalization (center), and specificity (right) of the model for the pose (top) and speaker subspace (bottom).
    For the generalization and specificity, we visualize the mean (red line) and the standard deviation (blue ribbon) of the experiments.
  }
  \label{fig:evaluation-both}
\end{figure}

\begin{figure}
  \centering
  \includegraphics{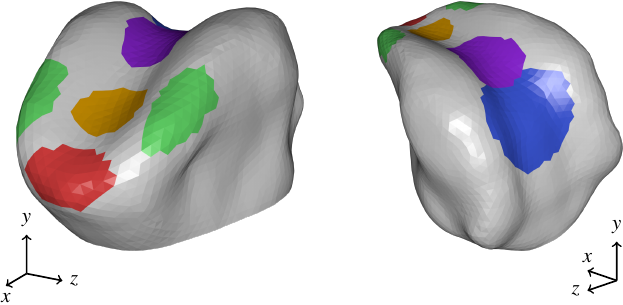}
  \caption{Speech related regions of the tongue surface:
    Tongue tip (red), tongue blade (brown), tongue back (violet), tongue dorsum (blue), and the lateral regions (green).
    }
  \label{fig:critical-regions}
\end{figure}

It is common to evaluate such statistical models by analyzing their compactness, generalization, and specificity \citep{Styner2003} in order to find the optimal subspace dimensionality.

\subsection{Compactness}

Compactness investigates how much the individual components of \(\vec{s}\) and \(\vec{p}\) contribute to the description of the used training data.
In \autoref{fig:evaluation-both}, we see that using \(\widetilde{m} = 5\) is sufficient to represent \SI{91}{\percent} of data variability.
Approximately the same holds for \(\widetilde{n} = 4\).

\subsection{Generalization}

Generalization measures how well the model can register data that was not part of the training.
To evaluate the speaker generalization, we designed the following experiment:
for each speaker, we derived a tongue model from the meshes of all other speakers and registered the tongue meshes of the excluded speaker.
Afterwards, we measured the average Euclidean distance between the registered meshes and the original ones.
Additionally, we analyzed the fitting results for different values of \(\widetilde{m}\).
The dimensionality of the pose subspace was fixed to \(\widetilde{n} = 4\) during these experiments to prevent overfitting caused by this subspace.

In the analysis of the pose generalization, we used a similar approach:
for each phone, we derived a tongue model from the meshes of all other phones and registered the tongue meshes of the excluded phone.
In this case, the dimensionality of the speaker subspace was fixed to \(\widetilde{m} = 5\).
The results of these experiments are depicted in \autoref{fig:evaluation-both}.
During this evaluation, we used the scale factor \(h = 2\) in the model fitting optimization.
We observe that increasing the subspace dimensionality leads to better fitting results.
The results of the generalization experiments show that only a few components of \(\vec{p}\) and \(\vec{s}\) are needed to reliably register unseen data, which implies that the model can adapt to new tongue anatomies or poses.
In particular, for \(\vec{p}\), 3 components are enough to reach an average error that is slightly above the average measurement resolution  of \SI{~1.2}{\mm} of the \ac{mri} scan data.
For \(\vec{s}\), 7 components are needed to reach this level of precision.
Furthermore, we observe that a high number of components leads to errors below the average measurement resolution of the scan data, which can be considered as overfitting.
We observe that the pose subspace has better generalization abilities than the speaker subspace.
We suspect this might be related to redundancies in our training data:
for example, the phone pairs [ʌ, ʉ], [e, i], and [e, ɛ] are similar to each other with respect to shape \citep{Ladefoged1982}.
This means that excluding one still provides the model with enough information to capture the related variation.

\subsection{Specificity}

Specificity tries to assess how much randomly generated tongue shapes of the model differ from valid tongue shapes.
This is essentially a measure for determining how specific the model is to the tongue.
In particular, we wanted to investigate how large these differences were for the regions of the tongue mesh that are speech related.
\autoref{fig:critical-regions} shows an overview of those regions.
To this end, we designed a few experiments where samples from the two subspaces were drawn randomly in order to generate a random tongue shape.
In these experiments, we used the tongue meshes of all speakers as the collection of valid shapes.
The first experiment investigated the specificity of the speaker subspace.
The pose subspace is again fixed to \(\widetilde{n} = 4\) and the speaker subspace size was varied.
For each value of \(\widetilde{m}\), we generated random tongue shapes and evaluated the average Euclidean distance between the created mesh and the closest one in the mesh collection.
In this comparison and distance evaluation, a region consisting of all speech related parts was considered.
The same experiment was conducted for analyzing the specificity of the pose subspace where the speaker subspace size was set to \(\widetilde{m} = 5\).
The results of both experiments can be inspected in \autoref{fig:evaluation-both}.
In these experiments, we see that increasing the subspace dimensionality leads to higher average Euclidean distances, which means that the model is becoming less specific.
This could also be seen as an indicator that the higher dimensions are modeling the noise in the training meshes.

\begin{figure}
  \centering
  \includegraphics{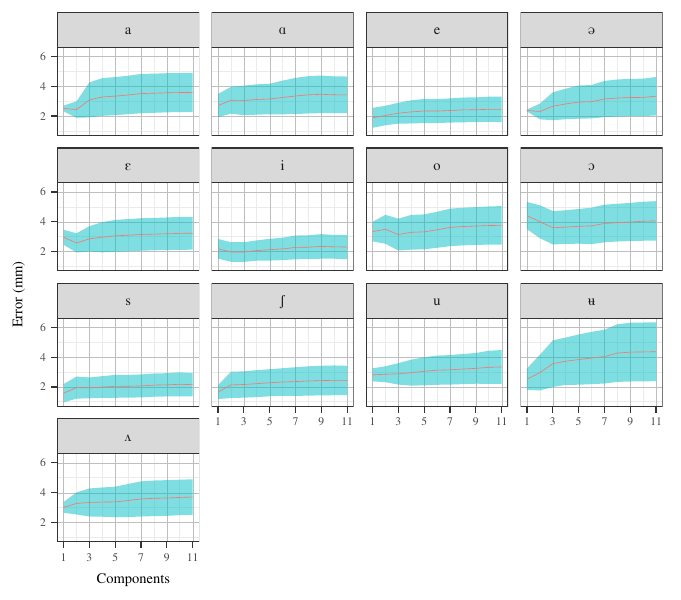}
  \caption{Specificity results for the individual fixed phone experiments.
    Plots show mean (red line) and standard deviation (blue ribbon).
  }
  \label{fig:fixed-phone-specificity}
\end{figure}

Finally, we wanted to find out how much the tongue shapes belonging to specific phones differ from the corresponding ones generated by the model.
We performed for each phone the following experiment:
we froze the coordinates in the pose subspace to the ones belonging to the given phone.
Moreover, we only allowed the generated meshes to be compared to meshes belonging to that phone.
Then, for each dimensionality of the speaker subspace, we generated samples and computed the average Euclidean distance to the closest mesh.
This time, in the distance evaluation and comparison, we used parts of the tongue that are considered critical for this specific phone \citep{Jackson2009}.
For the vowels [i, e, ɛ, a, ɑ, ʌ, ɔ, o, u, ʉ, ə], we selected a region consisting of the tongue blade, tongue back, and tongue dorsum.
The area for the sibilants [s, ʃ] contains the tongue tip and the tongue blade.
The results of these experiments are shown in \autoref{fig:fixed-phone-specificity}.
In all specificity experiments, we generated \num{1000000} samples.
In these experiments, we notice that the phone [ʉ] shows a significantly bad result in the fixed phone specificity evaluation, which might be related to its unusual role in the phonology of British English.
We suspect that some speakers might have pronounced it inconsistently and applied different strategies, which led to a high variation in the data, which is then integrated into the model.

Overall, we decided that setting \(\widetilde{m} = 5\) and \(\widetilde{n} = 4\) provides a good compromise between specificity, generalization, and compactness.
We note that this choice also limits the effects of overfitting.

\section{Using the model for tracking motion capture data}
\label{sec:tracking}

After having derived our final model with \(\widetilde{m} = 5\) and \(\widetilde{n} = 4\), we investigated if it could be used to reliably track the tongue motion capture data of an unknown speaker and to generate a plausible animation from it.
To this end, we decided to use \ac{ema} data from a previous study \citep{Steiner2014ISSP}.
This study focuses on the German language and consists of speakers that are not part of the Ultrax data.
\Ac{ema} uses alternating electromagnetic fields to track the position of coils that are attached to specific points of interest, \eg, on the tongue surface;
this modality can provide data with a high temporal resolution, but only gives access to a sparse set of points.

\subsection{Data selection and setup for experiments}

We selected the following data of the female subject \emph{VP05} in the dataset:
one recording that contains repeated consonant-vowel combinations of the consonants [f, s, ʃ, ç, x, ʁ, m, n, ŋ, l] and the vowel [u].
Furthermore, we used an \ac{ema} recording of the German translation of the ``Northwind and the Sun'' passage, a standard specimen in phonetic research \citep{ipahandbook1999}.
To combine this \ac{ema} data with our model, we had to register dynamic speech data of an unknown speaker containing new phonemes, which is a significant challenge.
However, we want to point out that the phonetic inventories of the English and German languages are very similar.
Thus, we think that the pose subspace of our tongue model should be compatible with motion capture data originating from producing German speech.

The raw \ac{ema} data was prepared as follows:
first, we smoothed the data to dampen any high-frequency measurement noise.
Afterwards, we removed rigid motion originating from head movements by using the three reference coils of the \ac{ema} data that were attached to suitable positions on the head.
Finally, we used the palate shape and the bite plane of the subject to rigidly align the data to our tongue model in a semi-supervised way.

For our experiments, we chose the 3 coils that were placed at the tongue tip, the tongue blade, and the tongue dorsum, which lay roughly in the mid-sagittal plane of the tongue.
We normalized this data, specifically, we projected the positional data into the mid-sagittal plane to guarantee this mid-sagittal property.

For our tracking experiments, we first had to find for each \ac{ema} coil a corresponding vertex on our tongue mesh.
We used the following semi-supervised approach to determine these correspondences from one frame of the used \ac{ema} data:
first, we sampled a random tongue shape from our model and initially determined for each coil the nearest neighbor on the mid-sagittal area of the tongue mesh.
Then, we iteratively refined these correspondences by fitting the model and updating the nearest neighbors.
We repeated the above two steps multiple times and kept the correspondences that achieved the smallest average distance between coil positions and their corresponding vertices.
Afterwards, we visually compared the proposed correspondences with a photographic reference of the subject's tongue with the attached coils and reran the above approach until the correspondences were plausible.
During the sampling and the correspondence optimization, we used the scale factor \(h = 1\) to avoid overfitting.

\subsection{Experiments}

In our experiments, we used the following energy to fit our model to the current \ac{ema} data frame:
\begin{align}
  E_{\text{Track}}(\vec{s}, \vec{p}) = \alpha\ E_\text{data}(\vec{s}, \vec{p}) + \beta\ E_\text{bias} (\vec{s}, \vec{p}) + \gamma\ E_\text{coherence}(\vec{s}, \vec{p})
  \label{eq:track-energy}
\end{align}

The data term \(E_{\text{data}}\) measures the distances between the vertex locations and their corresponding \ac{ema} coil positions.
The bias term \(E_{\text{bias}}\) penalizes deviations from the mean weights of the model.
We added this term to the energy to provide the approach with information about the average tongue shape in order to cope with the sparsity of the data.
Finally, the consistency term \(E_{\text{coherence}}\) favors a temporal coherence between consecutive frames.

For all experiments, we used \(\alpha = 1\) and \(\beta = \gamma = 5\) to provide a good compromise between these model ideas.
Furthermore, we set the scale factor \(h\) to \(5\) to give the approach sufficient freedom during the optimization.

In the first experiment, we optimized for \(\vec{s}\) and \(\vec{p}\).
However, we know that the anatomy of the speaker should remain constant during the recordings.
As it is unknown, we used a common approach to estimate the corresponding weights:
we averaged the obtained anatomy weights of the first experiment.
For the second experiment, we only optimized for \(\vec{p}\) and fixed \(\vec{s}\) to the estimated anatomy weights.

\begin{figure}
  \centering
  \includegraphics{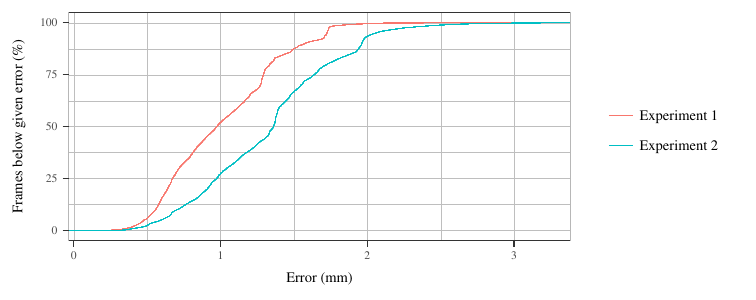}
  \caption{Cumulative error for the two tracking experiments.}
  \label{fig:tracking-errors}
\end{figure}

\begin{figure}
  \centering
  \includegraphics{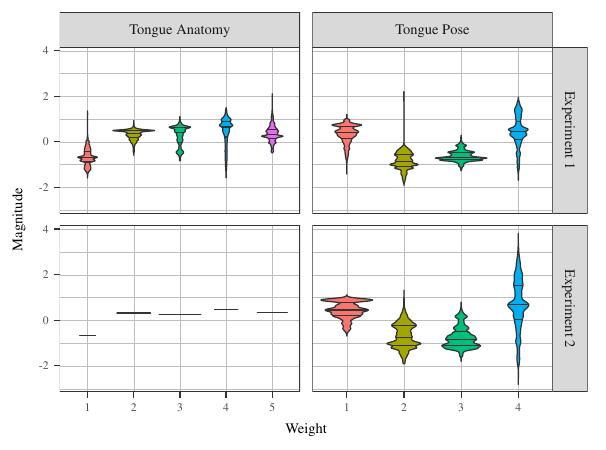}
  \caption{Weight distribution for the two tracking experiments.
    The violin plots show the density, with the mean and interquartile range marked by horizontal lines.
  }
  \label{fig:tracking-weights}
\end{figure}

\begin{table}
  \centering
  \smaller
  \input{build/assets/plotting/trackingStatistics_table.tex}
  \caption{Statistics of the weight distribution for the two tracking experiments.}
  \label{tab:tracking-weights}
\end{table}

For all \ac{ema} frames of the results, we computed the distribution of the weights and also the cumulative error.
The error in each frame was calculated by measuring the average Euclidean distance between vertex location and corresponding coil position.
The error is shown in \autoref{fig:tracking-errors} and the weight distributions can be inspected in \autoref{fig:tracking-weights} and \autoref{tab:tracking-weights}.
Additionally, we created for each experiment a video for the ``Northwind and the Sun'' passage.
These videos show the (anonymized) speaker during recording, an animation of the fitted tongue model together with the actual \ac{ema} coil positions, and information about the current weights.
Both videos can be found in the supplementary material:
the video file corresponding to the first experiment is named ``VP05\_full.mp4'', the second, ``VP05\_fixed\_anatomy.mp4''.
Note that we show normalized versions of the model weights, \ie, they are shifted and scaled such that \(x\) represents the value \(m_i + x\ \sigma_i\) where \(m_i\) and \(\sigma_i\) have the same roles as in \eqref{eq:value-limitation}.

\subsection{Discussion}

We observe in the results for the first experiment that we achieve acceptable errors where \SI{62}{\percent} of the errors are below \SI{1.25}{\mm}.
Additionally, we see that all weights are used during the tracking approach, which also means that the anatomy is often also adapted to improve the fitting result.
We see that their values approximately stay within the interval \(\left[m_i - 2\ \sigma_i, m_i + 2\ \sigma_i\right]\).
Moreover, we notice that weight 4 of the tongue pose is showing a significantly higher variance than the other pose weights.

Moving to the second experiment, we notice that the errors increased.
However, they are still acceptable:
\SI{62}{\percent} of the errors are still below \SI{1.5}{mm}.
This development can be seen as an expected behavior because the approach now only has 4 degrees of freedom instead of 9 to fit the data.
Moreover, we see that the variance of the tongue pose weights increased.
We might argue here that optimizing all weights as in the first experiment causes the pose weights to be underestimated.
Again, the fourth weight shows a higher variance than the others.
We suspect that this high variance and high range of achieved values might be caused by the presence of new phonemes or the fact that we are working with \ac{ema} data of German speakers.

By inspecting the video material, we notice that in the first experiment, the tongue is sometimes visually changing its anatomical properties:
for example, it is shrinking or expanding.
However, this is an expected behavior because the anatomy weights were also optimized to improve the result.
By fixing the anatomy and optimizing only the tongue pose in the second experiment, we seem to avoid these problems:
the anatomy of the tongue seems to stay stable and the transitions between frames also appear more plausible.
But we also discover in the video material that our model might be lacking important shape variations:
The motion of the tongue tip seems to be very rigid.

Overall, we conclude that the second approach produces more acceptable results than the first despite the slight loss of precision.
Moreover, we see that this approach also provides us with information about transition paths in the tongue pose subspace between phonemes.
These obtained transition paths could also be used to transfer the tracked motion to another speaker by adjusting the anatomy weights accordingly.

However, we observe that we cannot evaluate from these tracking experiments how close the estimated tongue shape is to the true tongue of the subject, because the true shape is unknown.
For such an evaluation, a dedicated study is needed, which assesses the tracking and reconstruction capabilities of the model; this is beyond the scope of the present work.
Such a evaluation might make use of an additional modality, such as \ac{us} or \ac{rtmri}, to determine the true tongue contour.

\section{Conclusion}
\label{sec:conclusion}

In this work, we have presented a multilinear tongue model that was derived from volumetric \ac{mri} scans in a minimally supervised way.
We also verified the validity of the extracted meshes by conducting a preference test.
Afterwards, we observed in the experiments that a model with a low dimensionality can reliably register unknown data with an acceptable precision.
Moreover, we explored in two experiments whether the model that was acquired from static data was suitable for tracking sparse motion capture data of the tongue.
We found that fixing the anatomical features of the model to a speaker specific shape provided the most acceptable results.
However, we also discovered indications that the current multilinear model might be missing some shape variations.

A previous version of our multilinear model was used in a text-to-speech framework that is able to synthesize audio with synchronized tongue motion \citep{Lemaguer2017IS}.
For this purpose, the audio and \ac{ema} recordings of the mngu0 articulatory database \citep{Richmond2011} were used, where we utilized the output tongue pose weights from a tracking approach similar to the one described in \autoref{sec:tracking} as feature vectors for training the synthesis framework.

In the future, we plan to investigate whether more shape variations can be obtained using more data.
To this end, we want to use additional datasets in our framework.
This implies that we also have to extract the shapes of phones like [ɡ, k] that are characterized by contact with the soft palate.
In this regard, we have to address the issue of recovering in the corresponding scans the surface of the soft palate, which can deform in a non-rigid way.
Additionally, the datasets we use might differ with respect to the recorded phones, which leads to missing data in our training set.
In this case, the simple averaging method for reconstructing missing shapes will no longer be sufficient.
We believe that such a model may have access to a sufficient amount of shape variations to be usable in areas of application such as audiovisual speech synthesis, \ac{capt}, or articulatory speech synthesis.
Of course, this hypothesis has to be carefully investigated and evaluated in future work.

Moreover, in the current study, we manually selected the parameters for our framework.
It may be interesting to investigate in the future if parameters exist that lead to acceptable results in a general setting, which would free the user from the burden of tuning them by hand. 
Moreover, we could try to improve the approach to extract higher quality meshes.

We are currently preparing a follow-up study to investigate the accuracy of the tongue shape reconstruction  from \ac{ema} data, by using \ac{rtmri} as a reference. 
In this context, we think it may also be worthwhile to explore whether the derived model can be used to extract realistic 3D tongue motion from 2D \ac{rtmri} data recorded in the mid-sagittal plane.
In contrast to \ac{ema}, this modality provides much richer motion information.

Finally, we could investigate if tongue motion transfer from one speaker to another is possible by adapting the anatomy weights to the new speaker and using the pose weights of the original speaker.
It would also be interesting to compare different types of dynamic speech, \eg, whispering, shouting, or expressive speech.
Ultimately, this could lead to a more flexible multilinear model that is able to synthesize these different types of speech.

\titlelabel{}

\section{Acknowledgments}

This study was funded by the German Research Foundation (grant number EXC 284).
It uses data from work supported by EPSRC Healthcare Partnerships (grant number EP/I027696/1).

\section{\refname}
\bibliographystyle{plainnat}
\urlstyle{same}
\bibliography{references}

\listoftodos

\end{document}

%% file: abstract.tex
We present a multilinear statistical model of the human tongue that captures anatomical and tongue pose related shape variations separately.
The model is derived from 3D magnetic resonance imaging data of 11 speakers sustaining speech related vocal tract configurations.
To extract model parameters, we use a minimally supervised method based on an image segmentation approach and a template fitting technique.
Furthermore, we use image denoising to deal with possibly corrupt data, palate surface information reconstruction to handle palatal tongue contacts, and a bootstrap strategy to refine the obtained shapes.
Our evaluation shows that, by limiting the degrees of freedom for the anatomical and speech related variations, to 5 and 4, respectively, we obtain a model that can reliably register unknown data while avoiding overfitting effects.
Furthermore, we show that it can be used to generate plausible tongue animation by tracking sparse motion capture data.

%% file: build/assets/plotting/trackingStatistics_table.tex
\sisetup{table-format=-1.2,round-mode=places,round-precision=2}
\begin{tabulary}{\linewidth}{CSSSSSSSS}
\toprule
& \multicolumn{4}{c}{Experiment 1} & \multicolumn{4}{c}{Experiment 2} \\
\cmidrule(lr){2-5} \cmidrule(lr){6-9}
{Weight} & {Mean} & {Std.\ dev.} & {Min} & {Max} & {Mean} & {Std.\ dev.} & {Min} & {Max} \\
\midrule
\(s_1\) & -0.6699 & 0.3643 & -1.5907 & 1.3494 & -0.6699 & 0 & -0.6699 & -0.6699 \\
\(s_2\) & 0.3163 & 0.2239 & -0.5921 & 0.9313 & 0.3163 & 0 & 0.3163 & 0.3163 \\
\(s_3\) & 0.256 & 0.4381 & -0.8361 & 1.1091 & 0.256 & 0 & 0.256 & 0.256 \\
\(s_4\) & 0.4577 & 0.6043 & -1.5818 & 1.4905 & 0.4577 & 0 & 0.4577 & 0.4577 \\
\(s_5\) & 0.3562 & 0.3441 & -0.484 & 2.1074 & 0.3562 & 0 & 0.3562 & 0.3562 \\
\(p_1\) & 0.3472 & 0.4021 & -1.4136 & 1.2008 & 0.4437 & 0.3679 & -0.6921 & 1.1123 \\
\(p_2\) & -0.8237 & 0.4208 & -1.9153 & 2.1971 & -0.6833 & 0.5366 & -1.9131 & 1.3214 \\
\(p_3\) & -0.6209 & 0.2354 & -1.2596 & 0.2855 & -0.7506 & 0.4591 & -1.8035 & 0.795 \\
\(p_4\) & 0.4509 & 0.6755 & -1.6906 & 1.9564 & 0.6953 & 1.107 & -2.8095 & 3.8136 \\
\bottomrule
\end{tabulary}

%% file: article.bbl
\begin{thebibliography}{89}
\providecommand{\natexlab}[1]{#1}
\providecommand{\url}[1]{\texttt{#1}}
\expandafter\ifx\csname urlstyle\endcsname\relax
  \providecommand{\doi}[1]{doi: #1}\else
  \providecommand{\doi}{doi: \begingroup \urlstyle{rm}\Url}\fi

\bibitem[Ult(2014)]{Ultrax2014}
Ultrax: Real-time tongue tracking for speech therapy using ultrasound, 2014.
\newblock URL \url{http://www.ultrax-speech.org/}.

\bibitem[Allen et~al.(2003)Allen, Curless, and Popovi\'{c}]{Allen2003}
Brett Allen, Brian Curless, and Zoran Popovi\'{c}.
\newblock The space of human body shapes: Reconstruction and parameterization
  from range scans.
\newblock \emph{ACM Transactions on Graphics}, 22\penalty0 (3):\penalty0
  587--594, 2003.
\newblock \doi{10.1145/1201775.882311}.

\bibitem[Ananthakrishnan et~al.(2010)Ananthakrishnan, Badin, Vargas, and
  Engwall]{Ananthakrishnan2010}
Gopal Ananthakrishnan, Pierre Badin, Juli{\'a}n Andr{\'e}s~Vald{\'e}s Vargas,
  and Olov Engwall.
\newblock Predicting unseen articulations from multi-speaker articulatory
  models.
\newblock In \emph{Interspeech}, pages 1588--1591, 2010.
\newblock URL
  \url{http://www.isca-speech.org/archive/interspeech_2010/i10_1588.html}.

\bibitem[Badin and Serrurier(2006)]{Badin2006}
Pierre Badin and Antoine Serrurier.
\newblock Three-dimensional linear modeling of tongue: Articulatory data and
  models.
\newblock In \emph{7th International Seminar on Speech Production (ISSP)},
  pages 395--402, 2006.

\bibitem[Badin et~al.(1998)Badin, Bailly, Raybaudi, and
  Segebarth]{badin1998three}
Pierre Badin, G{\'e}rard Bailly, Monica Raybaudi, and Christoph Segebarth.
\newblock A three-dimensional linear articulatory model based on {MRI} data.
\newblock In \emph{3rd ESCA/COCOSDA Workshop on Speech Synthesis (SSW)}, 1998.
\newblock URL \url{http://www.isca-speech.org/archive_open/ssw3/ssw3_249.html}.

\bibitem[Badin et~al.(2002)Badin, Bailly, Reveret, Baciu, Segebarth, and
  Savariaux]{badin2002three}
Pierre Badin, Gerard Bailly, Lionel Reveret, Monica Baciu, Christoph Segebarth,
  and Christophe Savariaux.
\newblock Three-dimensional linear articulatory modeling of tongue, lips and
  face, based on {MRI} and video images.
\newblock \emph{Journal of Phonetics}, 30\penalty0 (3):\penalty0 533--553,
  2002.
\newblock \doi{10.1006/jpho.2002.0166}.

\bibitem[Baer et~al.(1991)Baer, Gore, Gracco, and Nye]{baer1991analysis}
Thomas Baer, John~C. Gore, L.C. Gracco, and Patrick~W. Nye.
\newblock Analysis of vocal tract shape and dimensions using magnetic resonance
  imaging: Vowels.
\newblock \emph{Journal of the Acoustical Society of America}, 90\penalty0
  (2):\penalty0 799--828, 1991.
\newblock \doi{doi.org/10.1121/1.401949}.

\bibitem[Baker(2011)]{Baker2011}
Adam Baker.
\newblock A biomechanical tongue model for speech production based on {MRI}
  live speaker data, 2011.
\newblock URL \url{http://www.adambaker.org/qmu.php}.

\bibitem[Beautemps et~al.(2001)Beautemps, Badin, and
  Bailly]{beautemps2001linear}
Denis Beautemps, Pierre Badin, and G{\'e}rard Bailly.
\newblock Linear degrees of freedom in speech production: Analysis of
  cineradio- and labio-film data and articulatory-acoustic modeling.
\newblock \emph{Journal of the Acoustical Society of America}, 109\penalty0
  (5):\penalty0 2165--2180, 2001.
\newblock \doi{10.1121/1.1361090}.

\bibitem[Bijar et~al.(2016)Bijar, Rohan, Perrier, and Payan]{bijar2016atlas}
Ahmad Bijar, Pierre-Yves Rohan, Pascal Perrier, and Yohan Payan.
\newblock Atlas-based automatic generation of subject-specific finite element
  tongue meshes.
\newblock \emph{Annals of Biomedical Engineering}, 44\penalty0 (1):\penalty0
  16--34, 2016.
\newblock \doi{10.1007/s10439-015-1497-y}.

\bibitem[Blandin et~al.(2015)Blandin, Arnela, Laboissière, Pelorson, Guasch,
  Hirtum, and Laval]{Blandin2015JASA}
Rémi Blandin, Marc Arnela, Rafael Laboissière, Xavier Pelorson, Oriol Guasch,
  Annemie~Van Hirtum, and Xavier Laval.
\newblock Effects of higher order propagation modes in vocal tract like
  geometries.
\newblock \emph{Journal of the Acoustical Society of America}, 137\penalty0
  (2):\penalty0 832--843, 2015.
\newblock \doi{10.1121/1.4906166}.

\bibitem[Blanz and Vetter(1999)]{Blanz1999}
Volker Blanz and Thomas Vetter.
\newblock A morphable model for the synthesis of {3D} faces.
\newblock In \emph{26th Annual Conference on Computer Graphics and Interactive
  Techniques (SIGGRAPH)}, pages 187--194. ACM Press/Addison-Wesley Publishing
  Co., 1999.
\newblock \doi{10.1145/311535.311556}.

\bibitem[Bolkart and Wuhrer(2015)]{Bolkart2015}
Timo Bolkart and Stefanie Wuhrer.
\newblock {3D} faces in motion: Fully automatic registration and statistical
  analysis.
\newblock \emph{Computer Vision and Image Understanding}, 131:\penalty0
  100--115, 2015.
\newblock \doi{10.1016/j.cviu.2014.06.013}.

\bibitem[Botsch et~al.(2010)Botsch, Kobbelt, Pauly, Alliez, and
  Levy]{Botsch2010}
Mario Botsch, Leif Kobbelt, Mark Pauly, Pierre Alliez, and Bruno Levy.
\newblock \emph{Polygon Mesh Processing}.
\newblock A K Peters/CRC Press, 2010.

\bibitem[Brunner et~al.(2009)Brunner, Fuchs, and
  Perrier]{brunner2009relationship}
Jana Brunner, Susanne Fuchs, and Pascal Perrier.
\newblock On the relationship between palate shape and articulatory behavior.
\newblock \emph{Journal of the Acoustical Society of America}, 125\penalty0
  (6):\penalty0 3936--3949, 2009.
\newblock \doi{10.1121/1.3125313}.

\bibitem[Buchaillard et~al.(2007)Buchaillard, Brix, Perrier, and
  Payan]{buchaillard2007simulations}
Stéphanie Buchaillard, Muriel Brix, Pascal Perrier, and Yohan Payan.
\newblock Simulations of the consequences of tongue surgery on tongue mobility:
  Implications for speech production in post-surgery conditions.
\newblock \emph{International Journal of Medical Robotics and Computer Assisted
  Surgery}, 3\penalty0 (3):\penalty0 252--261, 2007.
\newblock \doi{10.1002/rcs.142}.

\bibitem[Buchaillard et~al.(2009)Buchaillard, Perrier, and
  Payan]{buchaillard2009biomechanical}
Stéphanie Buchaillard, Pascal Perrier, and Yohan Payan.
\newblock A biomechanical model of cardinal vowel production: Muscle
  activations and the impact of gravity on tongue positioning.
\newblock \emph{Journal of the Acoustical Society of America}, 126\penalty0
  (4):\penalty0 2033--2051, 2009.
\newblock \doi{10.1121/1.3204306}.

\bibitem[Burdumy et~al.(2015)Burdumy, Traser, Richter, Echternach, Korvink,
  Hennig, and Zaitsev]{burdumy2015acceleration}
Michael Burdumy, Louisa Traser, Bernhard Richter, Matthias Echternach, Jan~G
  Korvink, Jürgen Hennig, and Maxim Zaitsev.
\newblock Acceleration of {MRI} of the vocal tract provides additional insight
  into articulator modifications.
\newblock \emph{Journal of Magnetic Resonance Imaging}, 42\penalty0
  (4):\penalty0 925--935, 2015.
\newblock \doi{10.1002/jmri.24857}.

\bibitem[De~Silva and Lim(2008)]{deSilva2008}
Vin De~Silva and Lek-Heng Lim.
\newblock Tensor rank and the ill-posedness of the best low-rank approximation
  problem.
\newblock \emph{SIAM Journal on Matrix Analysis and Applications}, 30\penalty0
  (3):\penalty0 1084--1127, 2008.
\newblock \doi{10.1137/06066518X}.

\bibitem[Demolin et~al.(2000)Demolin, Metens, and Soquet]{demolin2000real}
Didier Demolin, Thierry Metens, and Alain Soquet.
\newblock Real time {MRI} and articulatory coordinations in vowels.
\newblock In \emph{5th Speech Production Seminar (SSP)}, pages 86--93, 2000.

\bibitem[Dryden and Mardia(1998)]{Dryden1998}
Ian~L Dryden and Kanti~V Mardia.
\newblock \emph{Statistical Shape Analysis}.
\newblock Wiley, 1998.

\bibitem[Elie et~al.(2016)Elie, Laprie, Vuissoz, and Odille]{elie2016}
Benjamin Elie, Yves Laprie, Pierre-André Vuissoz, and Freddy Odille.
\newblock High spatiotemporal {cineMRI} films using compressed sensing for
  acquiring articulatory data.
\newblock In \emph{24th European Signal Processing Conference (EUSIPCO)}, pages
  1353--1357, 2016.
\newblock \doi{10.1109/EUSIPCO.2016.7760469}.

\bibitem[Engwall(2000)]{Engwall2000}
Olov Engwall.
\newblock A {3D} tongue model based on {MRI} data.
\newblock In \emph{6th International Conference on Spoken Language Processing
  (ICSLP)}, volume~3, pages 901--904, 2000.
\newblock URL
  \url{http://www.isca-speech.org/archive/icslp_2000/i00_3901.html}.

\bibitem[Engwall(2008)]{Engwall2008}
Olov Engwall.
\newblock Can audio-visual instructions help learners improve their
  articulation? - an ultrasound study of short term changes.
\newblock In \emph{Interspeech}, pages 2631--2634, 2008.
\newblock URL
  \url{http://www.isca-speech.org/archive/interspeech_2008/i08_2631.html}.

\bibitem[Engwall and Badin(1999)]{engwall1999collecting}
Olov Engwall and Pierre Badin.
\newblock Collecting and analysing two- and three-dimensional {MRI} data for
  {S}wedish.
\newblock \emph{KTH Dept. for Speech, Music and Hearing Quarterly Progress and
  Status Report}, 40\penalty0 (3-4), 1999.
\newblock URL
  \url{http://www.speech.kth.se/prod/publications/files/qpsr/1999/1999_40_3-4_011-038.pdf}.

\bibitem[Eryildirim and Berger(2011)]{Eryildirim2011}
Abdulkadir Eryildirim and Marie-Odile Berger.
\newblock A guided approach for automatic segmentation and modeling of the
  vocal tract in {MRI} images.
\newblock In \emph{19th European Signal Processing Conference (EUSIPCO)}, pages
  61--65, 2011.
\newblock URL
  \url{http://www.eurasip.org/Proceedings/Eusipco/Eusipco2011/papers/1569425007.pdf}.

\bibitem[Fang et~al.(2016)Fang, Chen, Wang, Wei, Wang, Wu, and Li]{Fang2016}
Qiang Fang, Yun Chen, Haibo Wang, Jianguo Wei, Jianrong Wang, Xiyu Wu, and
  Aijun Li.
\newblock An improved {3D} geometric tongue model.
\newblock In \emph{Interspeech}, pages 1104--1107, 2016.
\newblock \doi{10.21437/Interspeech.2016-901}.

\bibitem[Foldvik(1993)]{foldvik1993time}
Arne~Kjell Foldvik.
\newblock A time-evolving three-dimensional vocal tract model by means of
  magnetic resonance imaging ({MRI}).
\newblock In \emph{3rd European Conference on Speech Communication and
  Technology (Eurospeech)}, pages 557--559, 1993.
\newblock URL
  \url{http://www.isca-speech.org/archive/eurospeech_1993/e93_0557.html}.

\bibitem[Fu et~al.(2015)Fu, Zhao, Carignan, Shosted, Perry, Kuehn, Liang, and
  Sutton]{fu2015high}
Maojing Fu, Bo~Zhao, Christopher Carignan, Ryan~K. Shosted, Jamie~L Perry,
  David~P. Kuehn, Zhi-Pei Liang, and Bradley~P. Sutton.
\newblock High-resolution dynamic speech imaging with joint low-rank and
  sparsity constraints.
\newblock \emph{Magnetic Resonance in Medicine}, 73\penalty0 (5):\penalty0
  1820--1832, 2015.
\newblock \doi{10.1002/mrm.25302}.

\bibitem[Fuchs et~al.(2008)Fuchs, Winkler, and Perrier]{fuchs2008speakers}
Susanne Fuchs, Ralf Winkler, and Pascal Perrier.
\newblock Do speakers' vocal tract geometries shape their articulatory vowel
  space?
\newblock In \emph{8th International Seminar on Speech Production (ISSP)},
  pages 333--336, 2008.
\newblock URL \url{http://issp2008.loria.fr/Proceedings/PDF/issp2008-77.pdf}.

\bibitem[Geng and Mooshammer(2009)]{geng2009stretch}
Christian Geng and Christine Mooshammer.
\newblock How to stretch and shrink vowel systems: Results from a vowel
  normalization procedure.
\newblock \emph{Journal of the Acoustical Society of America}, 125\penalty0
  (5):\penalty0 3278--3288, 2009.
\newblock \doi{10.1121/1.3106130}.

\bibitem[Harandi et~al.(2014)Harandi, Abugharbieh, and Fels]{Harandi2014Taylor}
Negar~M. Harandi, Rafeef Abugharbieh, and Sidney Fels.
\newblock {3D} segmentation of the tongue in {MRI}: a minimally interactive
  model-based approach.
\newblock \emph{Computer Methods in Biomechanics and Biomedical Engineering:
  Imaging \& Visualization}, 2014.
\newblock \doi{10.1080/21681163.2013.864958}.

\bibitem[Harandi et~al.(2017)Harandi, Woo, Stone, Abugharbieh, and
  Fels]{harandi2017}
Negar~M. Harandi, Jonghye Woo, Maureen Stone, Rafeef Abugharbieh, and Sidney
  Fels.
\newblock Variability in muscle activation of simple speech motions: A
  biomechanical modeling approach.
\newblock \emph{Journal of the Acoustical Society of America}, 141\penalty0
  (4):\penalty0 2579--2590, 2017.
\newblock \doi{10.1121/1.4978420}.

\bibitem[Harshman et~al.(1977)Harshman, Ladefoged, and Goldstein]{Harshman1977}
Richard Harshman, Peter Ladefoged, and Louis Goldstein.
\newblock Factor analysis of tongue shapes.
\newblock \emph{Journal of the Acoustical Society of America}, 62\penalty0
  (3):\penalty0 693--707, 1977.
\newblock \doi{10.1121/1.381581}.

\bibitem[Harshman(1970)]{Harshman1970}
Richard~A Harshman.
\newblock Foundations of the {PARAFAC} procedure: Models and conditions for an
  ``explanatory'' multi-modal factor analysis.
\newblock \emph{UCLA Working Papers in Phonetics}, 16, 1970.
\newblock URL \url{http://escholarship.org/uc/item/0410x385}.

\bibitem[Hewer et~al.(2014)Hewer, Steiner, and Wuhrer]{Hewer2014}
Alexander Hewer, Ingmar Steiner, and Stefanie Wuhrer.
\newblock A hybrid approach to {3D} tongue modeling from vocal tract {MRI}
  using unsupervised image segmentation and mesh deformation.
\newblock In \emph{Interspeech}, pages 418--421, 2014.
\newblock URL
  \url{http://www.isca-speech.org/archive/interspeech_2014/i14_0418.html}.

\bibitem[Hewer et~al.(2015)Hewer, Steiner, Bolkart, Wuhrer, and
  Richmond]{Hewer2015}
Alexander Hewer, Ingmar Steiner, Timo Bolkart, Stefanie Wuhrer, and Korin
  Richmond.
\newblock A statistical shape space model of the palate surface trained on {3D}
  {MRI} scans of the vocal tract.
\newblock In \emph{18th International Congress of Phonetic Sciences (ICPhS)},
  2015.
\newblock URL
  \url{https://www.internationalphoneticassociation.org/icphs-proceedings/ICPhS2015/Papers/ICPHS0724.pdf}.

\bibitem[Honda et~al.(1996)Honda, Maeda, Hashi, Dembowski, and
  Westbury]{honda1996human}
Kiyoshi Honda, Shinji Maeda, Michiko Hashi, Jim~S. Dembowski, and John~R.
  Westbury.
\newblock Human palate and related structures: their articulatory consequences.
\newblock In \emph{4th International Conference on Spoken Language Processing
  (ICSLP)}, pages 784--787. IEEE, 1996.
\newblock URL
  \url{http://www.isca-speech.org/archive/icslp_1996/i96_0784.html}.

\bibitem[Hoole et~al.(2000)Hoole, Wismüller, Leinsinger, Kroos, Geumann, and
  Inoue]{Hoole2000}
Phil Hoole, Axel Wismüller, Gerda Leinsinger, Christian Kroos, Anja Geumann,
  and Michiko Inoue.
\newblock Analysis of tongue configuration in multi-speaker, multi-volume {MRI}
  data.
\newblock In \emph{5th Seminar on Speech Production (SSP)}, pages 157--160,
  2000.

\bibitem[Hoole et~al.(2003)Hoole, Zierdt, and Geng]{Hoole2003}
Phil Hoole, Andreas Zierdt, and Christian Geng.
\newblock Beyond {2D} in articulatory data acquisition and analysis.
\newblock In \emph{15th International Congress of Phonetic Sciences (ICPhS)},
  pages 265--268, 2003.
\newblock URL
  \url{https://www.internationalphoneticassociation.org/icphs-proceedings/ICPhS2003/p15_0265.html}.

\bibitem[{International Phonetic Association}(1999)]{ipahandbook1999}
{International Phonetic Association}.
\newblock \emph{Handbook of the International Phonetic Association}.
\newblock Cambridge University Press, 1999.

\bibitem[Jackson and Singampalli(2009)]{Jackson2009}
Philip J.~B. Jackson and Veena~D. Singampalli.
\newblock Statistical identification of articulation constraints in the
  production of speech.
\newblock \emph{Speech Communication}, 51\penalty0 (8):\penalty0 695--710,
  2009.
\newblock \doi{10.1016/j.specom.2009.03.007}.

\bibitem[Johnson et~al.(1993)Johnson, Ladefoged, and
  Lindau]{johnson1993individual}
Keith Johnson, Peter Ladefoged, and Mona Lindau.
\newblock Individual differences in vowel production.
\newblock \emph{Journal of the Acoustical Society of America}, 94\penalty0
  (2):\penalty0 701--714, 1993.
\newblock \doi{10.1121/1.406887}.

\bibitem[Kaburagi(2015)]{kaburagi2015morphological}
Tokihiko Kaburagi.
\newblock Morphological and acoustic analysis of the vocal tract using a
  multi-speaker volumetric {MRI} dataset.
\newblock In \emph{Interspeech}, pages 379--383, 2015.
\newblock URL
  \url{http://www.isca-speech.org/archive/interspeech_2015/i15_0379.html}.

\bibitem[Kiers and Krijnen(1991)]{Kiers1991}
Henk A.~L. Kiers and Wim~P. Krijnen.
\newblock An efficient algorithm for {PARAFAC} of three-way data with large
  numbers of observation units.
\newblock \emph{Psychometrika}, 56\penalty0 (1):\penalty0 147--152, 1991.
\newblock \doi{10.1007/BF02294592}.

\bibitem[Kim et~al.(2009)Kim, Narayanan, and Nayak]{Kim2009}
Yoon-Chul Kim, Shrikanth~S. Narayanan, and Krishna~S. Nayak.
\newblock Accelerated three-dimensional upper airway {MRI} using compressed
  sensing.
\newblock \emph{Magnetic Resonance in Medicine}, 61\penalty0 (6):\penalty0
  1434--1440, 2009.
\newblock \doi{10.1002/mrm.21953}.

\bibitem[Kröger et~al.(2000)Kröger, Winkler, Mooshammer, and
  Pompino-Marschall]{kroger2000estimation}
Bernd~J Kröger, Ralf Winkler, Christine Mooshammer, and Bernd
  Pompino-Marschall.
\newblock Estimation of vocal tract area function from magnetic resonance
  imaging: Preliminary results.
\newblock In \emph{5th Seminar on Speech Production (SSP)}, pages 333--336,
  2000.

\bibitem[Ladefoged(1982)]{Ladefoged1982}
Peter Ladefoged.
\newblock \emph{A Course in Phonetics}.
\newblock Harcourt Brace Jovanovich, second edition, 1982.

\bibitem[Ladefoged and Broadbent(1957)]{ladefoged1957information}
Peter Ladefoged and Donald~Eric Broadbent.
\newblock Information conveyed by vowels.
\newblock \emph{Journal of the Acoustical Society of America}, 29\penalty0
  (1):\penalty0 98--104, 1957.
\newblock \doi{10.1121/1.1908694}.

\bibitem[{Le Maguer} et~al.(2017){Le Maguer}, Steiner, and
  Hewer]{Lemaguer2017IS}
Sébastien {Le Maguer}, Ingmar Steiner, and Alexander Hewer.
\newblock An {HMM}/{DNN} comparison for synchronized text-to-speech and tongue
  motion synthesis.
\newblock In \emph{Interspeech}, pages 239--243, 2017.
\newblock \doi{10.21437/Interspeech.2017-936}.

\bibitem[Lee et~al.(2013)Lee, Woo, Xing, Murano, Stone, and Prince]{Lee2013}
Junghoon Lee, Jonghye Woo, Fangxu Xing, Emi~Z. Murano, Maureen Stone, and
  Jerry~L Prince.
\newblock Semi-automatic segmentation of the tongue for {3D} motion analysis
  with dynamic {MRI}.
\newblock In \emph{10th IEEE International Symposium on Biomedical Imaging
  (ISBI)}, pages 1465--1468, 2013.
\newblock \doi{10.1109/ISBI.2013.6556811}.

\bibitem[Li et~al.(2009)Li, Adams, Guibas, and Pauly]{Li2009}
Hao Li, Bart Adams, Leonidas~J. Guibas, and Mark Pauly.
\newblock Robust single-view geometry and motion reconstruction.
\newblock \emph{ACM Transactions on Graphics}, 28\penalty0 (5):\penalty0
  175:1--175:10, 2009.
\newblock \doi{10.1145/1618452.1618521}.

\bibitem[Lingala et~al.(2016)Lingala, Toutios, Toger, Lim, Zhu, Kim, Vaz,
  Narayanan, and Nayak]{Lingala2016}
Sajan~G Lingala, Asterios Toutios, Johannes Toger, Yongwan Lim, Yinghua Zhu,
  Yoon-Chul Kim, Colin Vaz, Shrikanth~S. Narayanan, and Krishna~S. Nayak.
\newblock State-of-the-art {MRI} protocol for comprehensive assessment of vocal
  tract structure and function.
\newblock In \emph{Interspeech}, pages 475--479, 2016.
\newblock \doi{10.21437/Interspeech.2016-559}.

\bibitem[Lingala et~al.(2017)Lingala, Zhu, Kim, Toutios, Narayanan, and
  Nayak]{lingala2017fast}
Sajan~Goud Lingala, Yinghua Zhu, Yoon-Chul Kim, Asterios Toutios, Shrikanth
  Narayanan, and Krishna~S. Nayak.
\newblock A fast and flexible {MRI} system for the study of dynamic vocal tract
  shaping.
\newblock \emph{Magnetic Resonance in Medicine}, 77\penalty0 (1):\penalty0
  112--125, 2017.
\newblock \doi{10.1002/mrm.26090}.

\bibitem[Liu et~al.(2013)Liu, Musialski, Wonka, and Ye]{Liu2013}
Ji~Liu, Przemyslaw Musialski, Peter Wonka, and Jieping Ye.
\newblock Tensor completion for estimating missing values in visual data.
\newblock \emph{IEEE Transactions on Pattern Analysis and Machine
  Intelligence}, 35\penalty0 (1):\penalty0 208--220, 2013.
\newblock \doi{10.1109/TPAMI.2012.39}.

\bibitem[McGurk and MacDonald(1976)]{Mcgurk1976}
Harry McGurk and John MacDonald.
\newblock Hearing lips and seeing voices.
\newblock \emph{Nature}, 264:\penalty0 746--748, 1976.
\newblock \doi{10.1038/264746a0}.

\bibitem[Mermelstein(1973)]{mermelstein1973articulatory}
Paul Mermelstein.
\newblock Articulatory model for the study of speech production.
\newblock \emph{Journal of the Acoustical Society of America}, 53\penalty0
  (4):\penalty0 1070--1082, 1973.
\newblock \doi{10.1121/1.1913427}.

\bibitem[Narayanan et~al.(1995)Narayanan, Alwan, and
  Haker]{narayanan1995articulatory}
Shrikanth~S. Narayanan, Abeer~A. Alwan, and Katherine Haker.
\newblock An articulatory study of fricative consonants using magnetic
  resonance imaging.
\newblock \emph{Journal of the Acoustical Society of America}, 98\penalty0
  (3):\penalty0 1325--1347, 1995.
\newblock \doi{10.1121/1.413469}.

\bibitem[Narayanan et~al.(1997)Narayanan, Alwan, and
  Haker]{narayanan1997toward}
Shrikanth~S. Narayanan, Abeer~A. Alwan, and Katherine Haker.
\newblock Toward articulatory-acoustic models for liquid approximants based on
  {MRI} and {EPG} data. part {I}. the laterals.
\newblock \emph{Journal of the Acoustical Society of America}, 101\penalty0
  (2):\penalty0 1064--1077, 1997.
\newblock \doi{10.1121/1.418030}.

\bibitem[Narayanan et~al.(2004)Narayanan, Nayak, Lee, Sethy, and
  Byrd]{narayanan2004}
Shrikanth~S. Narayanan, Krishna Nayak, Sungbok Lee, Abhinav Sethy, and Dani
  Byrd.
\newblock An approach to real-time magnetic resonance imaging for speech
  production.
\newblock \emph{Journal of the Acoustical Society of America}, 115\penalty0
  (4):\penalty0 1771--1776, 2004.
\newblock \doi{10.1121/1.1652588}.

\bibitem[Niebergall et~al.(2013)Niebergall, Zhang, Kunay, Keydana, Job, Uecker,
  and Frahm]{niebergall2013real}
Aaron Niebergall, Shuo Zhang, Esther Kunay, Götz Keydana, Michael Job, Martin
  Uecker, and Jens Frahm.
\newblock Real-time {MRI} of speaking at a resolution of 33 ms: Undersampled
  radial {FLASH} with nonlinear inverse reconstruction.
\newblock \emph{Magnetic Resonance in Medicine}, 69\penalty0 (2):\penalty0
  477--485, 2013.
\newblock \doi{10.1002/mrm.24276}.

\bibitem[Otsu(1979)]{Otsu1975}
Nobuyuki Otsu.
\newblock A threshold selection method from gray-level histograms.
\newblock \emph{IEEE Transactions on Systems, Man, and Cybernetics}, 9\penalty0
  (1):\penalty0 62--66, 1979.
\newblock \doi{10.1109/TSMC.1979.4310076}.

\bibitem[Peng et~al.(2010)Peng, Kerrien, and Berger]{Peng2010}
Ting Peng, Erwan Kerrien, and Marie-Odile Berger.
\newblock A shape-based framework to segmentation of tongue contours from {MRI}
  data.
\newblock In \emph{IEEE International Conference on Acoustics, Speech and
  Signal Processing (ICASSP)}, pages 662--665, 2010.
\newblock \doi{10.1109/ICASSP.2010.5495123}.

\bibitem[Raeesy et~al.(2013)Raeesy, Rueda, Udupa, and Coleman]{Raeesy2013}
Zeynab Raeesy, Sylvia Rueda, Jayaram~K Udupa, and John Coleman.
\newblock Automatic segmentation of vocal tract {MR} images.
\newblock In \emph{10th IEEE International Symposium on Biomedical Imaging
  (ISBI)}, pages 1328--1331, 2013.
\newblock \doi{10.1109/ISBI.2013.6556777}.

\bibitem[Richmond et~al.(2011)Richmond, Hoole, and King]{Richmond2011}
Korin Richmond, Phil Hoole, and Simon King.
\newblock Announcing the electromagnetic articulography (day 1) subset of the
  mngu0 articulatory corpus.
\newblock In \emph{Interspeech}, pages 1505--1508, 2011.
\newblock URL
  \url{http://www.isca-speech.org/archive/interspeech_2011/i11_1505.html}.

\bibitem[Rodrigues et~al.(2001)Rodrigues, Gillies, and
  Charters]{rodrigues2001biomechanical}
Maria Andréia~Formico Rodrigues, Duncan~F. Gillies, and Peter Charters.
\newblock A biomechanical model of the upper airways for simulating
  laryngoscopy.
\newblock \emph{Computer Methods in Biomechanics and Biomedical Engineering},
  4\penalty0 (2):\penalty0 127--148, 2001.
\newblock \doi{10.1080/10255840008908001}.

\bibitem[Rosset et~al.(2004)Rosset, Spadola, and Ratib]{Rosset2004}
Antoine Rosset, Luca Spadola, and Osman Ratib.
\newblock {OsiriX}: an open-source software for navigating in multidimensional
  {DICOM} images.
\newblock \emph{Journal of Digital Imaging}, 17\penalty0 (3):\penalty0
  205--216, 2004.
\newblock \doi{10.1007/s10278-004-1014-6}.

\bibitem[Rudy and Yunusova(2013)]{rudy2013effect}
Krista Rudy and Yana Yunusova.
\newblock The effect of anatomic factors on tongue position variability during
  consonants.
\newblock \emph{Journal of Speech, Language, and Hearing Research}, 56\penalty0
  (1):\penalty0 137--149, 2013.
\newblock \doi{10.1044/1092-4388(2012/11-0218)}.

\bibitem[Scott et~al.(2012)Scott, Boubertakh, Birch, and Miquel]{scott2012}
A.D. Scott, R.~Boubertakh, M.J. Birch, and M.E. Miquel.
\newblock Towards clinical assessment of velopharyngeal closure using {MRI}:
  evaluation of real-time {MRI} sequences at 1.5 and 3 {T}.
\newblock \emph{British Journal of Radiology}, 85\penalty0 (1019):\penalty0
  e1083--e1092, 2012.
\newblock \doi{10.1259/bjr/32938996}.

\bibitem[Serrurier et~al.(2017)Serrurier, Badin, Boë, Lamalle, and
  Neuschaefer-Rube]{serrurier2017inter}
Antoine Serrurier, Pierre Badin, Louis-Jean Boë, Laurent Lamalle, and
  Christiane Neuschaefer-Rube.
\newblock Inter-speaker variability: speaker normalisation and quantitative
  estimation of articulatory invariants in speech production for {F}rench.
\newblock In \emph{Interspeech}, pages 2272--2276, 2017.
\newblock \doi{10.21437/Interspeech.2017-1126}.

\bibitem[Shadle et~al.(1999)Shadle, Mohammad, Carter, and Jackson]{shadle1999}
Christine~H. Shadle, Mohammad Mohammad, John~N. Carter, and Phillip J.~B.
  Jackson.
\newblock Multi-planar dynamic magnetic resonance imaging: New tools for speech
  research.
\newblock In \emph{14th International Congress of Phonetic Sciences (ICPhS)},
  pages 623--626, 1999.
\newblock URL
  \url{https://www.internationalphoneticassociation.org/icphs-proceedings/ICPhS1999/p14_0623.html}.

\bibitem[Steiner et~al.(2014)Steiner, Knopp, Musche, Schmiedel, Braun, and
  Ouni]{Steiner2014ISSP}
Ingmar Steiner, Peter Knopp, Sebastian Musche, Astrid Schmiedel, Angelika
  Braun, and Slim Ouni.
\newblock Investigating the effects of posture and noise on speech production.
\newblock In \emph{10th International Seminar on Speech Production (ISSP)},
  pages 417--420, 2014.

\bibitem[Stone and Lele(1992)]{stone1992representing}
Maureen Stone and Subhash Lele.
\newblock Representing the tongue surface with curve fits.
\newblock In \emph{2nd International Conference on Spoken Language Processing
  (ICSLP)}, pages 875--878, 1992.
\newblock URL
  \url{http://www.isca-speech.org/archive/icslp_1992/i92_0875.html}.

\bibitem[Stone and Lundberg(1996)]{stone1996three}
Maureen Stone and Andrew Lundberg.
\newblock Three-dimensional tongue surface shapes of {E}nglish consonants and
  vowels.
\newblock \emph{Journal of the Acoustical Society of America}, 99\penalty0
  (6):\penalty0 3728--3737, 1996.
\newblock \doi{10.1121/1.414969}.

\bibitem[Stone et~al.(2016)Stone, Woo, Lee, Poole, Seagraves, Chung, Kim,
  Murano, Prince, and Blemker]{stone2016structure}
Maureen Stone, Jonghye Woo, Junghoon Lee, Tera Poole, Amy Seagraves, Michael
  Chung, Eric Kim, Emi~Z. Murano, Jerry~L. Prince, and Silvia~S. Blemker.
\newblock Structure and variability in human tongue muscle anatomy.
\newblock \emph{Computer Methods in Biomechanics and Biomedical Engineering:
  Imaging \& Visualization}, pages 1--9, 2016.
\newblock \doi{10.1080/21681163.2016.1162752}.

\bibitem[Styner et~al.(2003)Styner, Rajamani, Nolte, Zsemlye, Székely, Taylor,
  and Davies]{Styner2003}
Martin~A. Styner, Kumar~T. Rajamani, Lutz-Peter Nolte, Gabriel Zsemlye, Gábor
  Székely, Christopher~J. Taylor, and Rhodri~H. Davies.
\newblock Evaluation of {3D} correspondence methods for model building.
\newblock In \emph{18th International Conference on Information Processing in
  Medical Imaging (IPMI)}, pages 63--75, 2003.
\newblock \doi{10.1007/978-3-540-45087-0_6}.

\bibitem[Tiede et~al.(1996)Tiede, Yehia, and
  Vatikiotis-Bateson]{tiede1996shape}
Mark~K. Tiede, Hani Yehia, and Eric Vatikiotis-Bateson.
\newblock A shape-based approach to vocal tract area function estimation.
\newblock In \emph{1st ETRW on Speech Production Modeling}, pages 41--44, 1996.
\newblock URL \url{http://www.isca-speech.org/archive/spm_96/sps6_041.html}.

\bibitem[Toutios and Narayanan(2015)]{toutios2015factor}
Asterios Toutios and Shrikanth~S. Narayanan.
\newblock Factor analysis of vocal-tract outlines derived from real-time
  magnetic resonance imaging data.
\newblock In \emph{18th International Congress of Phonetic Sciences (ICPhS)},
  2015.
\newblock URL
  \url{https://www.internationalphoneticassociation.org/icphs-proceedings/ICPhS2015/Papers/ICPHS0514.pdf}.

\bibitem[Tucker(1966)]{Tucker1966}
Ledyard~R Tucker.
\newblock Some mathematical notes on three-mode factor analysis.
\newblock \emph{Psychometrika}, 31\penalty0 (3):\penalty0 279--311, 1966.
\newblock \doi{10.1007/BF02289464}.

\bibitem[Valdés~Vargas et~al.(2012{\natexlab{a}})Valdés~Vargas, Badin,
  Ananthakrishnan, and Lamalle]{Vargas2012}
Julián~Andrés Valdés~Vargas, Pierre Badin, Gopal Ananthakrishnan, and
  Laurent Lamalle.
\newblock Articulatory speaker normalisation based on {MRI}-data using
  three-way linear decomposition methods.
\newblock In \emph{29e Journées d'Études sur la Parole (JEP)}, pages
  529--536, 2012{\natexlab{a}}.
\newblock URL \url{http://www.aclweb.org/anthology/F12-1067}.

\bibitem[Valdés~Vargas et~al.(2012{\natexlab{b}})Valdés~Vargas, Badin, and
  Lamalle]{Valdes2012}
Julián~Andrés Valdés~Vargas, Pierre Badin, and Laurent Lamalle.
\newblock Articulatory speaker normalisation based on {MRI}-data using
  three-way linear decomposition methods.
\newblock In \emph{Interspeech}, pages 2186--2189, 2012{\natexlab{b}}.
\newblock URL
  \url{http://www.isca-speech.org/archive/interspeech_2012/i12_2186.html}.

\bibitem[Weickert(1998)]{Weickert1998}
Joachim Weickert.
\newblock \emph{Anisotropic Diffusion in Image Processing}.
\newblock Teubner, 1998.

\bibitem[Weirich and Fuchs(2013)]{weirich2013palatal}
Melanie Weirich and Susanne Fuchs.
\newblock Palatal morphology can influence speaker-specific realizations of
  phonemic contrasts.
\newblock \emph{Journal of Speech, Language, and Hearing Research}, 56\penalty0
  (6):\penalty0 S1894--S1908, 2013.
\newblock \doi{10.1044/1092-4388(2013/12-0217)}.

\bibitem[Weirich et~al.(2013)Weirich, Lancia, and Brunner]{weirich2013inter}
Melanie Weirich, Leonardo Lancia, and Jana Brunner.
\newblock Inter-speaker articulatory variability during vowel-consonant-vowel
  sequences in twins and unrelated speakers.
\newblock \emph{Journal of the Acoustical Society of America}, 134\penalty0
  (5):\penalty0 3766--3780, 2013.
\newblock \doi{10.1121/1.4822480}.

\bibitem[Woo et~al.(2015{\natexlab{a}})Woo, Lee, Murano, Xing, Al-Talib, Stone,
  and Prince]{woo2015high}
Jonghye Woo, Junghoon Lee, Emi~Z. Murano, Fangxu Xing, Meena Al-Talib, Maureen
  Stone, and Jerry~L. Prince.
\newblock A high-resolution atlas and statistical model of the vocal tract from
  structural {MRI}.
\newblock \emph{Computer Methods in Biomechanics and Biomedical Engineering:
  Imaging \& Visualization}, 3\penalty0 (1):\penalty0 47--60,
  2015{\natexlab{a}}.
\newblock \doi{10.1080/21681163.2014.933679}.

\bibitem[Woo et~al.(2015{\natexlab{b}})Woo, Xing, Lee, Stone, and
  Prince]{woo2015construction}
Jonghye Woo, Fangxu Xing, Junghoon Lee, Maureen Stone, and Jerry~L. Prince.
\newblock Construction of an unbiased spatio-temporal atlas of the tongue
  during speech.
\newblock In \emph{24th International Conference on Information Processing in
  Medical Imaging (IPMI)}, pages 723--732, 2015{\natexlab{b}}.
\newblock \doi{10.1007/978-3-319-19992-4_57}.

\bibitem[Wu et~al.(2014)Wu, Dang, and Stavness]{wu2014iterative}
Xiyu Wu, Jianwu Dang, and Ian Stavness.
\newblock Iterative method to estimate muscle activation with a physiological
  articulatory model.
\newblock \emph{Acoustical Science and Technology}, 35\penalty0 (4):\penalty0
  201--212, 2014.
\newblock \doi{10.1250/ast.35.201}.

\bibitem[Yunusova et~al.(2012)Yunusova, Rosenthal, Rudy, Baljko, and
  Daskalogiannakis]{yunusova2012positional}
Yana Yunusova, Jeffrey~S Rosenthal, Krista Rudy, Melanie Baljko, and John
  Daskalogiannakis.
\newblock Positional targets for lingual consonants defined using
  electromagnetic articulography.
\newblock \emph{Journal of the Acoustical Society of America}, 132\penalty0
  (2):\penalty0 1027--1038, 2012.
\newblock \doi{10.1121/1.4733542}.

\bibitem[Zheng et~al.(2003)Zheng, Hasegawa-Johnson, and Pizza]{Zheng2003}
Yanli Zheng, Mark Hasegawa-Johnson, and Shamala Pizza.
\newblock Analysis of the three-dimensional tongue shape using a three-index
  factor analysis model.
\newblock \emph{Journal of the Acoustical Society of America}, 113\penalty0
  (1):\penalty0 478--486, 2003.
\newblock \doi{10.1121/1.1520538}.

\end{thebibliography}
